%% file: main.tex
\documentclass[letterpaper, 10 pt, twocolumn, conference, numbers,sort&compress]{ieeeconf} 

\IEEEoverridecommandlockouts                              

\usepackage{cite}
\usepackage{amsmath,amssymb,amsfonts}
\usepackage{hyperref}
\hypersetup{
    colorlinks = true,
    linkcolor = blue,
    urlcolor = blue,
}

\usepackage{graphicx}
\usepackage{booktabs}
\usepackage{caption}
\usepackage{textcomp}
\usepackage{amsfonts}       
\usepackage{algpseudocode}
\usepackage{algorithm}
\usepackage{xcolor}
\def\BibTeX{{\rm B\kern-.05em{\sc i\kern-.025em b}\kern-.08em
    T\kern-.1667em\lower.7ex\hbox{E}\kern-.125emX}}

\begin{document}

\newcommand{\model}{AutoJoin} 

\title{\LARGE \bf
\model: Efficient Adversarial Training against Gradient-Free Perturbations for Robust Maneuvering via Denoising Autoencoder and Joint Learning
}
\author{Michael Villarreal$^{1}$, Bibek Poudel$^{1}$, Ryan Wickman$^{2}$, Yu Shen$^{3}$, and Weizi Li$^{1}$   
\thanks{$^{1}$Michael Villarreal, Bibek Poudel, and Weizi Li are with Min H. Kao Department of Electrical Engineering and Computer Science at University of Tennessee, Knoxville, TN, USA {\tt\small \{tvillarr, bpoudel3\}@vols.utk.edu, weizili@utk.edu}}%
\thanks{$^{2}$Ryan Wickman is with Department of Computer Science at University of Memphis, TN, USA {\tt\small rwickman@memphis.edu}}
\thanks{$^{3}$Yu Shen is with Department of Computer Science at University of Maryland, College Park, MD, USA {\tt\small yushen@umd.edu}}
}

\maketitle

\begin{abstract}
With the growing use of machine learning algorithms and ubiquitous sensors, many `perception-to-control' systems are being developed and deployed. To ensure their trustworthiness, improving their robustness through adversarial training is one potential approach. 
We propose a gradient-free adversarial training technique, named \model{}, to effectively and efficiently produce robust models for image-based maneuvering. 
Compared to other state-of-the-art methods with testing on over 5M images, \model{} achieves significant performance increases up to the 40\% range against perturbations while improving on clean performance up to 300\%. 
\model{} is also highly efficient, saving up to 86\% time per training epoch and 90\% training data over other state-of-the-art techniques. 
The core idea of \model{} is to use a decoder attachment to the original regression model creating a denoising autoencoder within the architecture. 
This architecture allows the tasks `maneuvering' and `denoising sensor input' to be jointly learnt and reinforce each other's performance. 
\end{abstract}

\input{sections/intro}
\input{sections/related}
\input{sections/method}

\input{sections/experiments}
\input{sections/conclusion}
\bibliographystyle{IEEEtran}
\bibliography{bibliography}

\clearpage
\input{sections/appendix}

\end{document}

%% file: sections/intro.tex
\section{Introduction}


The wide adoption of machine learning algorithms and ubiquitous sensors have together resulted in numerous tightly-coupled `perception-to-control' systems being deployed in the wild. In order for these systems to be trustworthy, robustness is an integral characteristic to be considered in addition to their effectiveness. Adversarial training aims to increase the robustness of machine learning models by exposing them to perturbations that arise from artificial attacks~\cite{goodfellow2014explaining, madry2017towards} or natural disturbances~\cite{shen2021gradient}. In this work, we focus on the impact of these perturbations on image-based maneuvering and the design of efficient adversarial training for obtaining robust models. 
The test task is `maneuvering through a front-facing camera'--which represents one of the hardest perception-to-control tasks since the input images are taken from partially observable, nondeterministic, dynamic, and continuous environments. 

Inspired by the finding that model robustness can be improved through learning with simulated perturbations~\cite{bhagoji2018enhancing}, effective techniques such as AugMix~\cite{hendrycks2019augmix}, AugMax~\cite{wang2021augmax}, MaxUp~\cite{gong2021maxup}, and AdvBN~\cite{shu2020prepare} have been introduced for language modeling, and image-based classification and segmentation. The focus of these studies is not \textit{efficient adversarial training for robust maneuvering}. 
AugMix is less effective to gradient-based adversarial attacks due to the lack of sufficiently intense augmentations;
AugMax, based on AugMix, is less efficient because it uses a gradient-based adversarial training procedure, which is also a limitation of AdvBN.  
MaxUp requires multiple forward passes for a single data point to determine the most harmful perturbation, which increases computational costs.

Recent work by Shen et al.~\cite{shen2021gradient} represents the state-of-the-art, gradient-free adversarial training method for achieving robust maneuvering against image perturbations. Their technique adopts Fr\'{e}chet Inception Distance (FID)~\cite{heusel2017gans} to determine distinct intensity levels of the perturbations that minimize model performance. Afterwards, datasets of single perturbations are generated. 
Before each round of training, the dataset that can minimize model performance is selected and incorporated with the clean dataset for training. A fine-tuning step is also introduced to boost model performance on clean images. While effective, examining the perturbation parameter space via FID adds complexity to the approach and using distinct intensity levels limits the model generalizability and hence robust efficacy. 
The approach also requires generating large datasets (2.1M images)
prior to training, burdening computation and storage. 
Additional inefficiency and algorithmic complexity occur at training as the pre-round selection of datasets requires testing against perturbed datasets, resulting in vast data passing through the model. 

We aim to develop a \emph{gradient-free and efficient} adversarial training technique for robust maneuvering. 
Fig.~\ref{fig:pipeline} illustrates our approach, \model{}. We divide a steering angle prediction model into an encoder and a regression head. 
The encoder is attached by a decoder to form a denoising autoencoder (DAE). 
Using the DAE alongside the prediction model is motivated by the assumption that prediction on clean data is easier than on perturbed data.
The DAE and the prediction model are jointly learnt: when perturbed images are forward passed, the reconstruction loss is added with the regression loss, enabling the encoder to simultaneously improve on `maneuvering' and `denoising sensor input.' 
\model{} enjoys efficiency as the extra computational cost stems only from passing the intermediate features through the decoder. 
\model{} is also easier to implement than Shen et al.~\cite{shen2021gradient} as perturbations are randomly sampled within a moving range that is determined by linear curriculum learning~\cite{bengio2009curriculum}. 
The FID is used minimally to determine the maximum intensity of a perturbation. 
The model generalizability and robustness are significantly improved due to extensive exploration of the perturbation parameter space, and `denoising sensor input' provides the denoised training data for `maneuvering.'  

We test \model{} on four real-world driving datasets: Honda~\cite{ramanishka2018toward}, Waymo~\cite{sun2020scalability}, Audi~\cite{geyer2020a2d2}, and SullyChen~\cite{chen2017sully}, which total over 5M clean and perturbed images and show \model{} achieves \textbf{the best performance on the maneuvering task while being the most efficient.} For example, \model{} outperforms~\cite{shen2021gradient} up to \textbf{20\% in accuracy and 43\% in error reduction} using the Nvidia~\cite{bojarski2016end} backbone, and up to \textbf{44\% error reduction} compared to other adversarial training techniques when using the ResNet-50~\cite{he2016deep} backbone. \model{} is also highly efficient as it saves \textbf{8\% per epoch time} compared AugMix~\cite{hendrycks2019augmix} and saves \textbf{86\% per epoch time and 90\% training data} compared to Shen et al.~\cite{shen2021gradient}.   

We provide extensive ablation studies. For example, we find that using all perturbations (discussed in Sec.~\ref{sec:perturbation}) instead of a subset can avoid up to a 45\% accuracy reduction and 51\% error increase. We also find that not ensuring all perturbations are seen during learning and using distinct intensities from Shen et al.~\cite{shen2021gradient}, as opposed to random intensities, can cause up to a 16\% error increase. 
Furthermore, we observe that incorporating the denoised images generated by the DAE into the training process leads to a decrease in accuracy by up to 10\% and an increase in error by 42\%.
The project code and supplemental material is available at \href{https://github.com/Fluidic-City-Lab/AutoJoin}{\underline{https://github.com/Fluidic-City-Lab/AutoJoin}}

%% file: sections/related.tex
\section{Related Work}


Most adversarial training techniques against image perturbations to date have focused on image classification. For example, AugMix~\cite{hendrycks2019augmix} enhances model robustness and generalizability by layering randomly sampled augmentations together.
AugMax~\cite{wang2021augmax}, a derivation of AugMix, trains on AugMix-generated images and their gradient-based adversarial variants.
MaxUp~\cite{gong2021maxup} stochastically generates multiple augmented images of a single image and trains the model on the perturbed image that minimizes the model's performance.
As a result, MaxUp requires multiple passes of data through the model for determining the most harmful perturbation. 
AdvBN~\cite{shu2020prepare} is a gradient-based adversarial training technique that switches between batch normalization layers based on whether the training data is clean or perturbed. It achieves state-of-the-art performance when used with techniques such as AugMix on ImageNet-C~\cite{hendrycks2019benchmarking}. 


\begin{figure*}
    \centering
    \includegraphics[width=\linewidth]{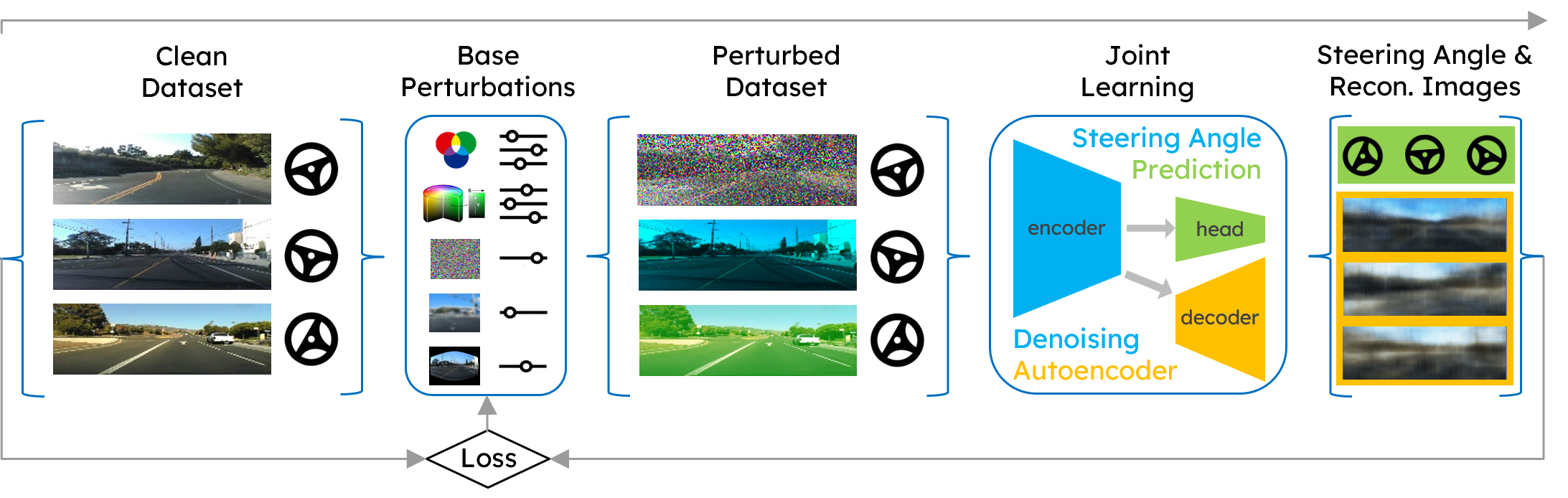} 
    \vspace{-1.5em}
    \caption{\small{The pipeline of \model{}. The clean data comes from real-world driving datasets containing front-facing camera images and their corresponding steering angles. 
    The perturbed data is prepared using various base perturbations and their sampled intensity levels. 
    The steering angle prediction model and denoising autoencoder are jointly learnt to reinforce each other's performance. 
    The resulting predictions and reconstructed images are used to compute the loss for adjusting perturbation intensity levels during learning. 
    }}
    \label{fig:pipeline}
    \vspace{-1em}
\end{figure*}

Recently, Shen et al.~\cite{shen2021gradient} has developed a gradient-free adversarial training technique against image perturbations. Their work uses Fr\'{e}chet Inception Distance (FID)~\cite{heusel2017gans} to select distinct intensity levels of perturbations. 
During training, the intensity that minimizes the current model's performance is adopted. While being the state-of-the-art method, the algorithmic pipeline combined with pre-training dataset generation are inefficient.
First, an extensive analysis is needed to determine five intensity levels of perturbations.
Second, the data selection process during training requires testing various combinations of perturbations and their distinct intensity levels. 
Third, significant costs are required for generating the pre-training datasets. 
In contrast, \model{} uses minimal FID analysis to obtain one point instead of five, which is then used for a range of intensities instead of having distinct intensities. \model{} also discards the mid-training data selection process in favor of ensuring all perturbations are seen each epoch. Lastly, \model{} generates perturbed datasets online during training.


DAEs have been used to improve model robustness for driving~\cite{roy2018robust, xiong2022robust, aspandi2019robust}. 
For example, Wang et al.~\cite{wang2020end} use an autoencoder to improve the accuracy of steering angle prediction by removing various roadside distractions such as trees or bushes. 
Their focus is not robustness against perturbed images as only clean images are used in training. 
DriveGuard~\cite{papachristodoulou2021driveguard} explores different autoencoder architectures on adversarially degraded images that affect semantic segmentation rather than the steering task. They show that autoencoders can be used to enhance the quality of the degraded images, thus improving overall task performance. 
Xie et al.~\cite{xie2019feature} and Liao et al.~\cite{liao2018defense} use denoising as a method component to improve on their tasks' performance, where the focus is gradient-based attacks~\cite{chen2020adversarial}, rather than gradient-free perturbations. 
The tasks are also restricted to classification instead of regression. 
Studies by Hendrycks et al.~\cite{hendrycks2019using} and Chen et al.~\cite{chen2020adversarial} adopt self-supervised training to improve model robustness. 
However, their focus is again on (image) classification and not regression. 
To the best of our knowledge, our work is \emph{the first gradient-free and efficient adversarial training technique for improving model robustness against perturbed image input in driving}.



%% file: sections/method.tex
\section{Methodology}

The pipeline of \model{} in shown in Fig.~\ref{fig:pipeline}. We use four driving datasets Honda~\cite{ramanishka2018toward}, Waymo~\cite{sun2020scalability}, A2D2~\cite{geyer2020a2d2}, and SullyChen~\cite{chen2017sully}) in training and evaluation.
During training, each image is perturbed by selecting a perturbation from a pre-determined set at a sampled intensity level (Sec.~\ref{sec:perturbation}). 
The perturbed images are then passed through DAE and the regression head for joint learning (Sec.~\ref{sec:joint_learning}). 

\subsection{Image Perturbations and Intensity Levels} 
\label{sec:perturbation}
We use the same base perturbations as of Shen et al.~\cite{shen2021gradient} for fair comparisons. 
We first perturb images' RGB color values and HSV saturation/brightness values in two directions, lighter or darker according to $v'_c = \alpha(a_c || b_c) + (1-\alpha)v_c$, where $v'_c$ is the perturbed pixel value, $\alpha$ is the intensity level, $a_c$ is the channel value's lower bound, $b_c$ is the channel value's upper bound, and $v_c$ is the original pixel value. $a_c$ is used for the darker direction and has the default value 0 and $b_c$ is used for the lighter direction and has the default value 255.
Two exceptions exist: $a_c$ is set to 10 for the V channel to exclude a black image, and $b_c$ is set to 179 for the H channel according to its definition. 
The other base perturbations include Gaussian noise, Gaussian blur, and radial distortion, which are used to simulate natural corruptions to an image. The Gaussian noise and Gaussian blur are parameterized by the standard deviation of the image. Sample images of each base perturbation are shown in Appendix~\ref{app:exp}. 

In addition to the nine base perturbations, the channel perturbations (i.e., R, G, B, H, S, V) are further discretized into their lighter or darker components such that if $p$ is a channel perturbation, it is decomposed into $p_{light}$ and $p_{dark}$. As a result, the perturbation set contains 15 elements. During learning, we expose the model to all 15 perturbations with the aim to improve its generalizability and robustness. We refer to using the \textbf{F}ull \textbf{S}et of 15 perturbations as FS in Sec.~\ref{sec:results}.
\model{} is trained on images with single perturbations, yet proves effective not only on such images but also on those with multiple and unseen perturbations. 

The intensity level of a perturbation is sampled within $[0,c)$. The minimum 0 represents no perturbation, and $c$ is the current maximum intensity. The range is upper-bounded by $c_{max}$, whose value is inherited from Shen et al.~\cite{shen2021gradient} to ensure comparable experiments. In practice, we scale $[0,c_{max})$ to $[0,1)$. After each epoch of training, $c$ is increased by 0.1 providing the model loss has reduced comparing to previous epochs. The entire training process begins on clean images ($[0,0)$). In contrast to Shen et al., our approach allows the model to explore the entire parameter space of a perturbation (rather than on distinct intensity levels). We refer to using \textbf{R}andom \textbf{I}ntensities as RI in Sec.~\ref{sec:results}. Further exploration of the perturbation parameter space by altering the minimum/maximum values is discussed in Appendix~\ref{sec:app_minmaxintensity}. 


\subsection{Joint Learning} 
\label{sec:joint_learning}

\begin{algorithm}[tb]
\caption{\model{}}
\begin{algorithmic}
\State{\textbf{input: } training batch $\{x_{i}\}_n$ (clean images), encoder $e$, decoder $d$, regression model $p$, perturbations $\mathcal{M}$, curriculum bound $c$} 
\For{\textbf{each }epoch}
    \For{\textbf{each }$i \in$ {1,...,$n$}} 
        \State Select perturbation $op$ = $\mathcal{M}$[$i$ mod $len(\mathcal{M})$]
        \State Sample intensity level $l$ from [0, $c$) randomly
        \State $y_{i}$ = $op$($x_{i}$, $l$) // perturb a clean image
        \State $z_{i} = e(y_{i})$ // obtain the latent representation
        \State $x'_{i} = d(z_{i})$ // reconstruct an image from the latent representation
        \State $a_{p} = p(z_{i})$ // predict a steering angle using the latent representation
        \If{$i$ \% $len(\mathcal{M})$ = 0}
            \State Shuffle $\mathcal{M}$ // randomize the order of perturbations
        \EndIf
    \EndFor
    \State Calculate $\mathcal{L}$ using Eq.~\ref{loss_function}
    \If{$\mathcal{L}$ improves}
        \State Increase $c$ by 0.1 // increase the curriculum's difficulty
    \EndIf
    \State Update $e$, $d$, and $p$ to minimize $\mathcal{L}$
\EndFor
\State \textbf{return} $e$ and $p$ // for steering angle prediction  

\end{algorithmic}
\label{alg:autojoin}
\end{algorithm}

The denoising autoencoder (DAE) and steering angle prediction model are jointly learnt. 
The DAE learns how to denoise the perturbed sensor input, while the prediction model learns how to maneuver given the denoised input. Both models train the shared encoder's latent representations, resulting in positive transfer between the tasks for two reasons. First, the DAE trains the latent representations to be the denoised versions of perturbed images, which enables the regression head to be trained on denoised representations rather than noisy representations, which may deteriorate the task performance. Second, the prediction model trains the encoder's representations for better task performance, and since the DAE uses these representations, the reconstructions are improved in favoring the overall task.

Our approach is described in Algorithm~\ref{alg:autojoin}. For a clean image $x_i$, a perturbation and its intensity $l \in [0,c)$ are sampled. The augmented image $y_i$ is a function of the two and is passed through the encoder $e(\cdot)$ to obtain the latent representation $z_i$. Next, $z_i$ is passed through both the decoder $d(\cdot)$ and the regression model $p(\cdot)$, where the results are the reconstruction $x'_i$ and steering angle prediction $a_{p_{i}}$, respectively. We randomize the perturbation set every 15 images to prevent overfitting.

For the DAE, the standard $\ell_2$ loss is used by comparing $x'_i$ to $x_i$. For the regression loss, $\ell_1$ is used between $a_{p_{i}}$ and $a_{t_{i}}$, where the latter is the ground truth angle. The two losses are combined for the joint learning:


\vspace{-0.5em}
\begin{equation}
    \mathcal{L} = \lambda_1\ell_2 \left(\mathbf{x'_i},\mathbf{x_i}\right)+\lambda_2 \ell_1 \left(\mathbf{a_{p_{i}}},\mathbf{a_{t_{i}}}\right).
\label{loss_function}
\end{equation}

The weights $\lambda_1$ and $\lambda_2$ are set as follows. For the experiments on the Waymo~\cite{sun2020scalability} dataset, $\lambda_1$ is set to 10 and $\lambda_2$ is set to 1 for better performance (emphasizing reconstructions). For the other three datasets, $\lambda_1$ is set to 1 and $\lambda_2$ is set to 10 to ensure the main focus of the joint learning is `maneuvering.' Once training is finished, the decoder is detached, leaving the prediction model for testing through datasets in six categories (see Sec.~\ref{sec:exp_setup} for details).

%% file: sections/experiments.tex
\section{Experiments and Results}
\label{sec:results}



\subsection{Experiment Setup} 
\label{sec:exp_setup}

We compare \model{} to five other approaches: Shen et al.~\cite{shen2021gradient} (referred to as Shen hereafter), AugMix~\cite{hendrycks2019augmix}, MaxUp~\cite{gong2021maxup}, AdvBN~\cite{shu2020prepare}, and AugMax~\cite{wang2021augmax}. 
We also compare to a Standard model, one trained using only clean images, and a Standard model trained with the \textbf{F}ull \textbf{S}et of 15 perturbations and \textbf{R}andom \textbf{I}ntensities discussed in Sec.~\ref{sec:perturbation}. 
We test on two backbones, the Nvidia model~\cite{bojarski2016end} and ResNet-50~\cite{he2016deep}. The breakdown of the two backbones is detailed in Appendix~\ref{app:exp}. 
We use four driving datasets in our experiments: Honda~\cite{ramanishka2018toward}, Waymo~\cite{sun2020scalability}, A2D2~\cite{geyer2020a2d2}, and SullyChen~\cite{chen2017sully}.
They have been widely adopted for developing machine learning models for driving-related tasks~\cite{xu2019temporal,shi2020pv,yi2021complete,shen2021gradient}.
Based on these four datasets, we generate test datasets according to Shen to ensure fair comparisons. The test datasets contain more than 5M images in four categories: Clean, Single, Combined, and Unseen. Combined contains images each with several single perturbations overlaid and Unseen contains unobserved images with perturbations extracted from the ImageNet-C dataset~\cite{hendrycks2019benchmarking}. 
Sample images for each category are given in Fig.~\ref{fig:sample_test2}.
The details of these datasets and more sample images are given in Appendix~\ref{app:exp}. 

We evaluate our approach using mean accuracy (MA) and mean absolute error (MAE). MA is defined as 

\vspace{-1em}
\begin{equation}
    \sum_{\tau}acc_{\tau\in\mathcal{T}}/|\mathcal{T}|, \,\, acc_{\tau} = count(|a_{p} - a_{t}| < \tau)/n,
\end{equation}

\noindent where $n$ denotes the number of test cases, $\mathcal{T} =$ \{1.5, 3.0, 7.5, 15.0, 30.0, 75.0\}, and $a_{p}$ and $a_{t}$ are the predicted angle and true angle, respectively. Note that we do not use the AMAI/MMAI metrics, which are derived from MA scores, from Shen since AMAI/MMAI only show performance improvement while the actual MA scores are more comprehensive.
All experiments are conducted using Intel Core i7-11700k CPU with 32G RAM and Nvidia RTX 3080 GPU.
We use the Adam optimizer~\cite{kingma2014adam}, batch size 124, and learning rate $10^{-4}$ for training. All models are trained for 500 epochs.


\begin{figure*}
    \centering
    \includegraphics[width=.96\linewidth]{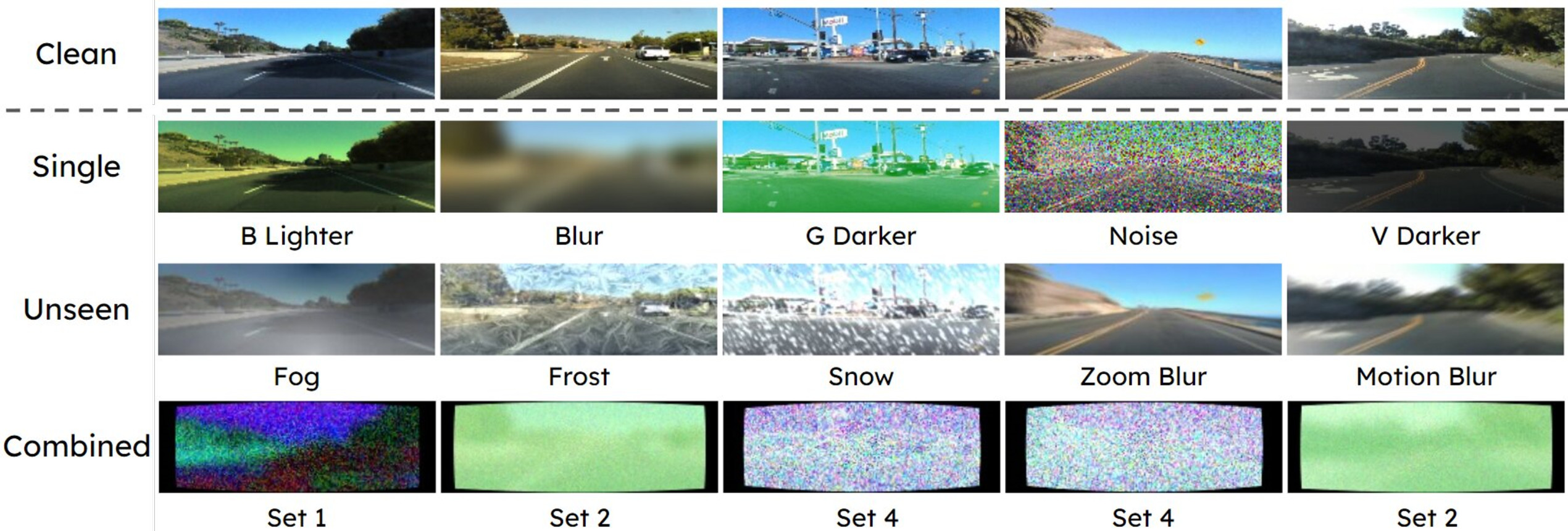}
    \caption{\small{Sample perturbed images. 
    Single is perturbed by only one of the perturbations outlined in Sec.~\ref{sec:perturbation} Unseen contains corruptions from ImageNet-C~\cite{hendrycks2019benchmarking}. Single and Unseen are selected with intensities from 0.5 to 1.0 to highlight the perturbation. Combined images have multiple perturbations overlaid, e.g., Set 2 includes G, noise, and blur as the most prominent perturbations. 
    }}
    \label{fig:sample_test2}
    \vspace{-1em}
\end{figure*}

\subsection{Results}
The main results, components of \model{}, and efficiency analysis are discussed in Sec.~\ref{sec:effectiveness-f}, Sec.~\ref{sec:design}, and Sec.~\ref{sec:efficiency}, respectively.  
The results reported are the averages over all test cases of a given test category.

\subsubsection{Effectiveness Against Gradient-free Perturbations}
\label{sec:effectiveness-f}

\begin{table}
    \caption{\small{Results on the SullyChen dataset using the Nvidia backbone. 
    \model{} outperforms all other techniques in all test categories and improves the clean performance three times over Shen when compared to Standard.}}
  \centering
  \addtolength{\tabcolsep}{-0.2em}
  \scalebox{0.77}{
  \begin{tabular}{lcccccccc}
    \toprule
     & \multicolumn{2}{c}{Clean} & \multicolumn{2}{c}{Single}  & \multicolumn{2}{c}{Combined} & \multicolumn{2}{c}{Unseen} \\
    \midrule
     & MA (\%) & MAE & MA (\%) & MAE & MA (\%) & MAE & MA (\%) & MAE \\
    \midrule
    Standard & 86.19 & 3.35 & 66.19 & 11.33 & 38.50 & 25.03 & 67.38 & 10.94 \\
    Standard (FSRI) & 78.21 & 4.73 & 74.91 & 5.97 & 61.54 & 11.93 & 71.98 & 7.24 \\
    AdvBN & 79.51 & 5.06 & 69.07 & 9.18  & 44.89 & 20.36 & 67.97 & 9.78  \\
    AugMix & 86.24 & 3.21 & 79.46 & 5.21 & 49.94 & 17.24 & 74.73 & 7.10  \\
    AugMax & 85.31 & 3.43 & 81.23 & 4.58 & 51.50 & 17.25 & 76.45 & 6.35 \\
    MaxUp & 79.15 & 4.40 & 77.40 & 5.01 & 61.72 & 12.21 & 73.46 & 6.71  \\
    Shen & 87.35 & 3.08 & 84.71 & 3.76 & 53.74 & 16.27 & 78.49 & 6.01 \\
    \midrule
    \model{} & \textbf{89.46} & \textbf{2.86} & \textbf{86.90} & \textbf{3.53} & \textbf{64.67} & \textbf{11.21} & \textbf{81.86} & \textbf{5.12} \\
    \bottomrule
  \end{tabular}}
\label{results_all_sully_nvidia}
\end{table}

Table~\ref{results_all_sully_nvidia} shows the comparison results on the SullyChen dataset using the Nvidia backbone. 
\model{} \emph{outperforms every other adversarial technique across all test categories} in both performance metrics. 
In particular, \model{} improves accuracy on Clean by 3.3\% MA and 0.58 MAE compared to the standard model trained solely on clean data. 
This result is significant as the clean performance is the most difficult to improve while \model{} achieves about \textbf{three times the improvement} on Clean compared to Shen. Tested on the perturbed datasets, \model{} achieves 64.67\% MA on Combined -- a \textbf{20\% accuracy increase} compared to Shen, 11.21 MAE on Combined -- a \textbf{31\% error decrease} compared to Shen, and 5.12 MAE on Unseen -- another \textbf{15\% error decrease} compared to Shen.  

Table~\ref{results_all_audi_nvidia} shows the comparison results on the A2D2 dataset using the Nvidia backbone. 
\model{} again \emph{outperforms all other techniques}. 
To list a few notable improvements over Shen: 6.7\% MA improvement on Clean to the standard model, which is a \textbf{4.2\% performance increase}; 11.72\% MA improvement -- a \textbf{17\% accuracy increase}, and 6.48 MAE drop -- a \textbf{43\% error decrease} on Combined; 5.01\% MA improvement -- a \textbf{7\% accuracy increase} and 1.76 MAE drop -- a \textbf{8\% error decrease} on Unseen. 

\begin{table}
  \caption{\small{Results on the A2D2 dataset using the Nvidia backbone. 
  \model{} outperforms every other approach in all test categories while improving on Clean by a wide margin of 4.2\% MA compared to Shen, achieving 168\% performance increase}.}
  \centering
  \addtolength{\tabcolsep}{-0.2em}
  \scalebox{0.77}{
  \begin{tabular}{lcccccccc}
    \toprule
     & \multicolumn{2}{c}{Clean} & \multicolumn{2}{c}{Single}  & \multicolumn{2}{c}{Combined} & \multicolumn{2}{c}{Unseen} \\
    \midrule
     & MA (\%) & MAE & MA (\%) & MAE & MA (\%) & MAE & MA (\%) & MAE \\
    \midrule
    Standard & 78.00 & 8.07 & 61.51 & 21.42 & 43.05 & 28.55 & 59.41 & 26.72 \\
    Standard (FSRI) & 77.95 & 8.33 & 77.31 & 8.64 & 72.45 & 10.39 & 74.04 & 10.49 \\
    AdvBN & 76.59 & 8.56 & 67.58 & 12.41 & 43.75 & 24.27 & 70.64 & 11.76  \\
    AugMix & 78.04 & 8.16 & 73.94 & 10.02 & 58.22 & 20.66 & 71.54 & 11.44  \\
    AugMax & 77.21 & 8.79 & 75.14 & 10.43 & 60.81 & 23.87 & 72.74 & 11.87 \\
    MaxUp & 78.93 & 8.17 & 78.36 & 8.42 & 71.56 & 13.22 & 76.78 & 9.24  \\
    Shen & 80.50 & 7.74 & 78.84 & 8.32 & 67.40 & 15.06 & 75.30 & 9.99 \\
    \midrule
    \model{} & \textbf{84.70} & \textbf{6.79} & \textbf{83.70} & \textbf{7.07} & \textbf{79.12} & \textbf{8.58} & \textbf{80.31} & \textbf{8.23} \\
    \bottomrule
  \end{tabular}} 
\label{results_all_audi_nvidia}
\vspace{-1em}
\end{table}

Switching to the ResNet-50 backbone, Table~\ref{results_honda_rn50} shows the results on the Honda dataset. 
Here, we only compare to Shen and AugMix because Shen is the latest technique and AugMix was the state-of-the-art before Shen, which also has the ability to improve both clean and robust performance on driving datasets.
As a result, \model{} \emph{outperforms both Shen and AugMix on perturbed datasets in most categories}. 
Specifically, \model{} achieves \textbf{the highest MAs across all perturbed categories}. 
\model{} also drops the MAE to 1.98 on Single, achieving \textbf{44\% improvement} over AugMix and \textbf{21\% improvement} over Shen; and drops the MAE to 2.89 on Unseen, achieving \textbf{33\% improvement} over AugMix and \textbf{41\% improvement} over Shen.  On this particular dataset, Shen outperforms \model{} on Clean by small margins due to its additional fine-tuning step on Clean. Nevertheless, \model{} still manages to improve upon the standard model and AugMix on Clean by large margins. 

\begin{table}
  \caption{\small{Results of comparing \model{} to AugMix and Shen on the Honda dataset using ResNet-50. 
  \model{} achieves the best overall robust performance. However, Shen's fine-tuning stage solely on Clean images grants them an advantage on Clean.}  
}
  \centering
  \addtolength{\tabcolsep}{-0.2em}
  \scalebox{0.77}{
  \begin{tabular}{lcccccccc}
    \toprule
     & \multicolumn{2}{c}{Clean} & \multicolumn{2}{c}{Single}  & \multicolumn{2}{c}{Combined} & \multicolumn{2}{c}{Unseen} \\
    \midrule
     & MA (\%) & MAE & MA (\%) & MAE & MA (\%) & MAE & MA (\%) & MAE  \\
    \midrule
    Standard & 92.87 & 1.63 & 73.12 & 11.86 & 55.01 & 22.73 & 69.92 & 13.65 \\
    Standard (FSRI) & 88.58 & 2.27 & 86.11 & 3.30 & 47.85 & 39.12 & 81.93 & 4.92 \\
    AugMix & 90.57 & 1.97 & 86.82 & 3.53 & 64.01 & 15.32 & 84.34 & 4.31 \\
    Shen & \textbf{97.07} & \textbf{0.93} & 93.08 & 2.52 & 70.53 & \textbf{13.20} & 87.91 & 4.94 \\
    \midrule
    \model{} & 96.46 & 1.12 & \textbf{94.58} & \textbf{1.98} & \textbf{70.70} & 14.56 & \textbf{91.92} & \textbf{2.89}  \\
    \bottomrule
  \end{tabular}}
\label{results_honda_rn50}
\end{table}

\begin{table}
  \caption{\small{Results of comparing our approaches (\model{} and \model{}-Fuse) to AugMix and Shen on the Waymo dataset using ResNet-50. 
  Our approaches not only improve on Clean the most, but also achieve the best overall robust performance.
}} 
  \centering
  \addtolength{\tabcolsep}{-0.2em}
  \scalebox{0.77}{
  \begin{tabular}{lcccccccc}
    \toprule
     & \multicolumn{2}{c}{Clean} & \multicolumn{2}{c}{Single} & \multicolumn{2}{c}{Combined} & \multicolumn{2}{c}{Unseen} \\
    \midrule
     & MA (\%) & MAE & MA (\%) & MAE & MA (\%) & MAE & MA (\%) & MAE \\
    \midrule
    Standard & 61.83 & 19.53 & 55.99 & 31.78 & 45.66 & 55.81 & 57.74 & 24.22 \\
    Standard (FSRI) & 61.83 & 20.15 & 61.45 & 20.29 & 56.95 & 24.94 & 60.56 & 21.06 \\
    AugMix & 61.74 & 19.19 & 60.83 & 20.10 & 56.34 & 24.23 & 59.78 & 21.75 \\
    Shen & 64.77 & 18.01 & 64.07 & 19.77 & 61.67 & \textbf{20.28} & 63.93 & 18.77 \\
    \midrule
    \model{} & 64.91 & 18.02 & 63.84 & 19.30 & 58.74 & 26.42 & 64.17 & 19.10 \\
    \model{}-Fuse & \textbf{65.07} & \textbf{17.60} & \textbf{64.34} & \textbf{18.49} & \textbf{63.48} & 20.82 & \textbf{65.01} & \textbf{18.17}  \\
    \bottomrule
  \end{tabular}}
\label{results_waymo_rn50}
\vspace{-2em}
\end{table}

During testing, we find Waymo to be unique in that the model benefits more from learning the inner representations of the denoised images. Therefore, we slightly modify the procedure of Algorithm~\ref{alg:autojoin} after perturbing the batch as follows: 1) one-tenth of the perturbed batch is sampled; 2) for each single perturbed image sampled, two other perturbed images are sampled; and 3) the three images are averaged to form a `fused' image. This is different from AugMix as AugMix applies multiple perturbations to a single image. We term this alternative procedure \model{}-Fuse. 

Table~\ref{results_waymo_rn50} shows the results on the Waymo dataset using ResNet-50. \model{}-Fuse makes a prominent impact by \emph{outperforming Shen on every test category except for combined MAE}. 
We also improve the clean performance over the standard model by \textbf{3.24\% MA and 1.93 MAE}. 
\model{} also outperforms AugMix by margins up to \textbf{7.14\% MA and 3.41 MAE}. 
These results are significant as for all four datasets, the well-performing robust techniques operate within 1\% MA or 1 MAE. 
While not the focus of this project, we additionally explore \model{}'s effectiveness against gradient-based adversarial examples. The results and discussion are provided in Appendix~\ref{app_gradient-based}

\subsubsection{Effectiveness of \model{} Pipeline}
\label{sec:design}


\begin{table}
  \caption{\small{Results of comparing \model{} with or without DAE on the SullyChen dataset with the Nvidia backbone. 
  Using DAE allows \model{} to achieve three times the performance gain on Clean over Shen. \model{} without DAE performs worse than Shen on Clean and Single MAE. 
  Ours without FSRI performs worse than Shen, but better than AugMix. These changes result in our method performing worse than Shen, showing the necessity of \model{}'s pipeline design.}} 
  \centering
  \addtolength{\tabcolsep}{-0.2em}
  \scalebox{0.77}{
  \begin{tabular}{lcccccccc}
    \toprule
     & \multicolumn{2}{c}{Clean} & \multicolumn{2}{c}{Single} & \multicolumn{2}{c}{Combined} & \multicolumn{2}{c}{Unseen} \\
    \midrule
     & MA (\%) & MAE & MA (\%) & MAE & MA (\%) & MAE & MA (\%) & MAE \\
    \midrule
    Standard & 86.19 & 3.35 & 66.19 & 11.33 & 38.50 & 25.03 & 67.38 & 10.94 \\
    AugMix & 86.24 & 3.21 & 79.46 & 5.21 & 49.94 & 17.24 & 74.73 & 7.10  \\
    Shen & 87.35 & 3.08 & 84.71 & 3.76 & 53.74 & 16.27 & 78.49 & 6.01 \\
    \midrule
    Ours, w/o DAE & 88.30 & 3.09 & 85.75 & 3.81 & 62.96 & 11.90 & 81.09 & 5.33 \\
    Ours, w/o FSRI & 86.43 & 3.54 & 83.19 & 4.62 & 61.97 & 13.01 & 78.51 & 6.23  \\
    Ours (\model{}) & \textbf{89.46} & \textbf{2.86} & \textbf{86.90} & \textbf{3.53} & \textbf{64.67} & \textbf{11.21} & \textbf{81.86} & \textbf{5.12} \\ 
    \bottomrule
  \end{tabular}}
\label{results_no_dae_sully} 
\vspace{-.5em}
\end{table}



\textbf{DAE and Feedback Loop}. A major component of \model{} is DAE. The results of our approach with or without DAE are shown in Table~\ref{results_no_dae_sully}. \model{} without DAE outperforms Shen in several test categories but not on Clean and Single MAE, meaning the perturbations and sampled intensity levels are effective for performance gains. 
In addition, a byproduct of DAE is denoised images. A natural idea is to use these images as additional training data for the prediction model, thus forming a feedback loop within \model{}. We explore the feedback loop in greater detail in Appendix~\ref{sec:app_feedback_loop}. 
Overall, we find a decrease in performance due to the feedback loop, thus we exclude it from the \model{}'s pipeline. 




\textbf{Perturbations and Intensities}. 
To better understand the base perturbations, we conduct experiments with six different perturbation subsets: 1) No RGB perturbations, 2) No HSV perturbations, 3) No blur, Gaussian noise, or distort perturbations (denoted as BND), 4) only RGB perturbations and Gaussian noise, 5) only HSV perturbations and Gaussian noise, and 6) no blur or distort perturbations. 
These subsets are formed to examine the effects of the color spaces and/or blur, distort, or Gaussian noise. 
We exclude Single from the results, shown in Table~\ref{ablation_perturb_sully_main} as different subsets will cause Single becoming a mixture of unseen and seen perturbations. 
Not using blur or distort outperforms using all perturbations within Combined by 1.11 MA and 0.33 MAE, but not within Clean and Unseen. 
We observe that using RGB perturbations tends to result in better Clean performance. Not using the HSV perturbations results in the worse generalization performance out of the models with 50.22\% MA and 78.91\% MA in Combined and Unseen, respectively. Overall, we find that using all 15 perturbations is necessary for maximal performance. More results and findings for the other driving datasets is given in Appendix~\ref{sec:app_perturb_study}.

We also assess \model{}'s performance without using the \textbf{F}ull \textbf{S}et of 15 perturbations and \textbf{R}andom \textbf{I}ntensities (FSRI), which are described as \model{} without FSRI in Table~\ref{results_no_dae_sully}. In this case, perturbations are randomly sampled during the learning process, and Shen's distinct intensity values are adopted. The results show a significant decrease in performance when FSRI are excluded. Thus, both FS and RI are necessary for optimal performance and to surpass Shen. More results and details on the FS and RI components for different datasets are provided in Appendix~\ref{sec:app_nine_static}.


\begin{table}
    \caption{\small{Results on the SullyChen dataset with the Nvidia backbone using six subsets of the base perturbations. `w/o BND' means no presence of blur, noise, and distort. Single is removed for a fair comparison. Using all base perturbations results in the best overall performance.}}
  \centering
  \scalebox{0.82}{
  \begin{tabular}{lcccccc}
    \toprule
     & \multicolumn{2}{c}{Clean} & \multicolumn{2}{c}{Combined} & \multicolumn{2}{c}{Unseen} \\
    \midrule
     & MA (\%) & MAE & MA (\%) & MAE & MA (\%) & MAE \\
    \midrule
    w/o RGB & 87.71 & 3.00 & 58.88 & 14.71 & 81.23 & 5.13 \\
    w/o HSV & 88.33 & 2.91 & 50.22 & 18.05 & 78.91 & 6.04  \\
    w/o BND & 88.24 & 3.05 & 59.49 & 12.80 & 80.45 & 5.55  \\
    RGB & 88.24 & 3.13 & 44.41 & 23.05 & 78.31 & 6.48 \\
    HSV & 88.66 & 3.04 & 54.78 & 15.35 & 80.60 & 5.60 \\
    RGB + Gaussian noise & 88.39 & 3.15 & 65.05 & 11.25 & 80.17 & 5.75 \\
    HSV + Gaussian noise & 86.70 & 3.52 & 63.34 & 11.82 & 79.49 & 5.87  \\
    w/o Blur \& Distort & 87.29 & 3.56 & \textbf{65.78} & \textbf{10.88} & 80.43 & 5.57 \\
    \midrule
    All & \textbf{89.46} & \textbf{2.86} & 64.67 & 11.21 & \textbf{81.86} & \textbf{5.12} \\
    \bottomrule
  \end{tabular}}
\label{ablation_perturb_sully_main}
\vspace{-1.0em}
\end{table}


\subsubsection{Efficiency}
\label{sec:efficiency}
We use AugMix/Shen + the Honda/Waymo datasets + ResNet-50 as the baselines for testing the efficiency.
On the Honda dataset, \model{} takes 109 seconds per epoch on average, while AugMix takes 118 seconds and Shen takes 759 seconds. Our approach \textbf{saves 8\% and 86\% per epoch time} compared to AugMix and Shen, respectively. 
On the Waymo dataset, AugMix takes 128 seconds and Shen takes 818 seconds, while \model{} takes only 118 seconds -- \textbf{8\% and 86\% time reduction} compared to AugMix and Shen, respectively. Note that the time listed for Shen excludes perturbation selection process during training, which adds to overall training time.
More time comparisons such as AugMix/Shen on the Sully/A2D2 datasets with the Nvidia backbone can be found in Appendix~\ref{sec:app_time_tables}.  
Lastly, our approach requires \textbf{90\% less training data} needed by Shen: we perturb the original clean dataset during the training process, unlike Shen's method of creating nine perturbed datasets (one for each base perturbation) beforehand and combining them with the clean dataset.

%% file: sections/conclusion.tex
\section{Conclusion, Limitations, and Future Work} 
\label{sec:conclusion}

We propose \model{}, a gradient-free adversarial training technique that is simple yet effective and efficient for robust maneuvering. We show that \model{} outperforms state-of-the-art adversarial techniques on various real-word driving datasets through extensive experimentation. \model{} is the most efficient technique tested on by being faster per epoch compared to AugMix and saving 83\% per epoch time and 90\% training data over Shen

AutoJoin is constrained to the regression task of autonomous driving rather than being a general-purpose data augmentation technique. Our focus is on autonomous driving given the significance in making autonomous driving systems robust to perturbations considering the real-world impacts (such as accidents or fatalities) that could occur if such driving systems fail. Another limitation is that while we use real-world images for training/testing, we have not deployed AutoJoin in a real-world test driving scenario. This limits our understanding of how AutoJoin would work within day-to-day driving scenarios. Such testing would be absolutely necessary in order to ensure the efficacy of AutoJoin in autonomous driving systems and the safety of all parties during the testing scenario.

In the future, we are interested in expanding \model{} to explore wider perturbation space and more intensity levels to remove any use of FID as well as using other perturbation sets. Furthermore, we would like to explore means to improve clean and combined performance on the Honda/Waymo datasets with ResNet-50. 
Although our work lacks theoretical support, this remains an open problem since the state-of-the-art techniques we compared with also lack theoretical evidence. Seeking theoretical support for our findings would be an interesting research direction.


\section*{Acknowledgement}
This research is supported by NSF IIS-2153426. The authors would like to thank NVIDIA and the Tickle College of Engineering at UTK for the support. 



%% file: sections/appendix.tex
\begin{appendices}

\section{Datasets and Experiment Setup}
\label{app:exp} 

\begin{figure*}
    \centering
    \includegraphics[width=\linewidth]{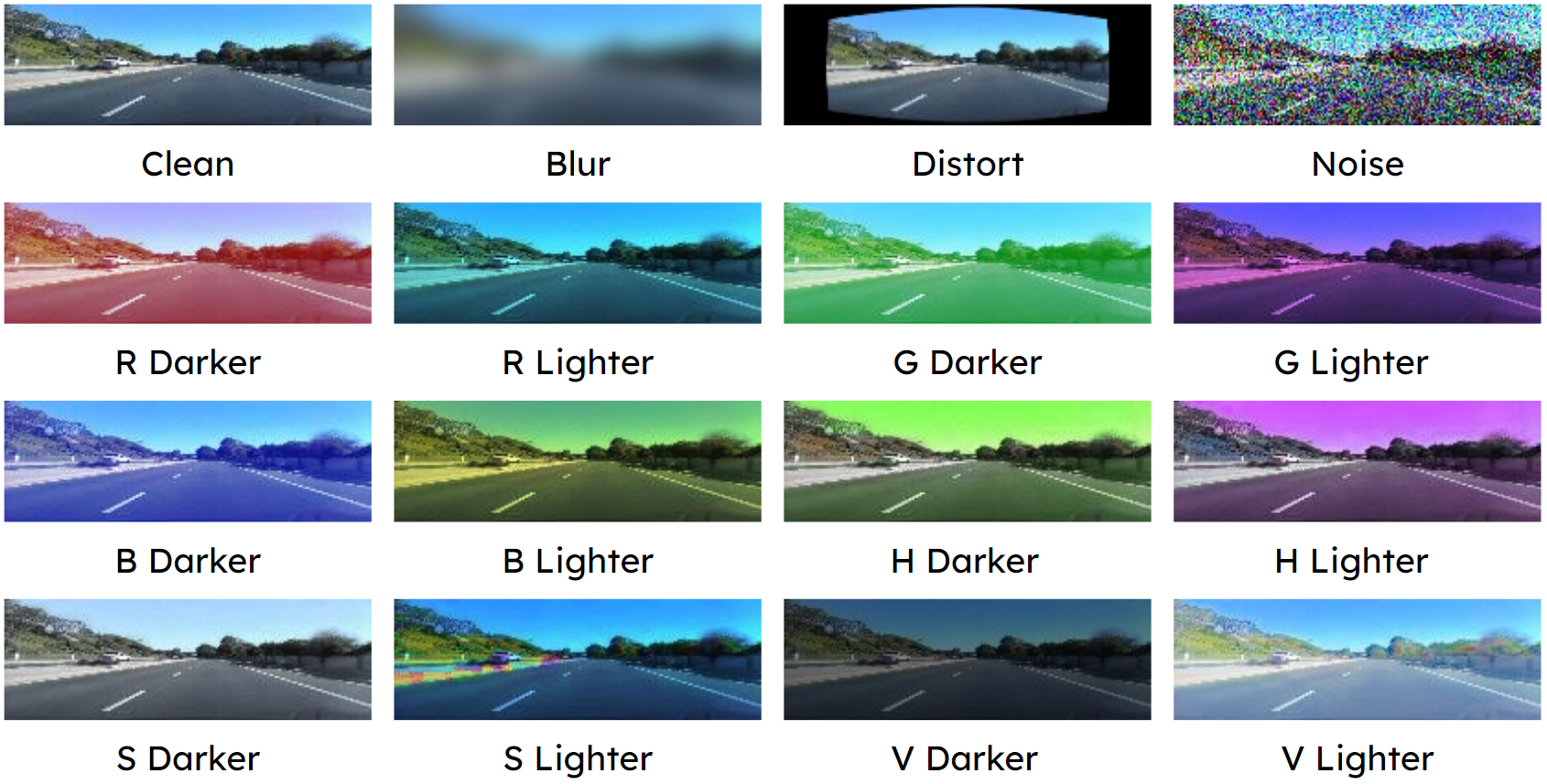}
    \caption{
    \small{
    Sample images used during training within the SullyChen~\cite{chen2017sully} dataset. The clean image and its perturbed variants using all base perturbations are shown. The intensity level of the images is 0.5, half of the max intensity. 
    }
    }
    \label{fig:sample_base}
\end{figure*}

\textbf{Base Perturbations.}
The description of the base perturbations is given in Sec.~\ref{sec:perturbation}. As an example, in Fig.~\ref{fig:sample_base} we show the clean image and the perturbed images from all base perturbations. The perturbation intensity is 0.5, half of the maximum intensity.

\textbf{Driving datasets and perturbed datasets.} 
We use four driving datasets in our experiments: Honda~\cite{ramanishka2018toward}, Waymo~\cite{sun2020scalability}, A2D2~\cite{geyer2020a2d2}, and SullyChen~\cite{chen2017sully}.
Theses datasets have been widely adopted for developing machine learning models for driving-related tasks~\cite{xu2019temporal,shi2020pv,yi2021complete,shen2021gradient}.
Based on these four datasets, we generate test datasets that contain more than 5M images in six categories. 
Four of them are gradient-free, named Clean, Single, Combined, Unseen, and are produced according to Shen to ensure fair comparisons. We also present details for two gradient-based datasets, FGSM and PGD, which are used to test our approach's adversarial transferability in Appendix~\ref{app_gradient-based}. 

\begin{itemize}
    \item Clean: the original driving datasets Honda, Waymo, A2D2, and SullyChen.
    \item Single: images with a single perturbation applied at five intensity levels from Shen over the 15 perturbations introduced in Sec.~\ref{sec:perturbation}. This results in 75 test cases in total. 
    \item Combined: images with multiple perturbations at the intensity levels drawn from Shen. There are six combined test cases in total.
    \item Unseen: images perturbed with simulated effects, including fog, snow, rain, frost, motion blur, zoom blur, and compression, from ImageNet-C~\cite{hendrycks2019benchmarking}. Each effect is perturbed at five intensity levels for a total of 35 unseen test cases.
    \item FGSM: adversarial images generated using FGSM~\cite{goodfellow2014explaining} with either the Nvidia model or ResNet-50 trained only on clean data. FGSM generates adversarial examples in a single step by maximizing the gradient of the loss function with respect to the images. We generate test cases within the bound $L_{\infty}$ norm at five step sizes $\epsilon = 0.01, 0.025, 0.05, 0.075$ and $0.1$. 
    \item PGD: adversarial images generated using PGD~\cite{madry2017towards} with either the Nvidia model or ResNet-50 trained only on clean data. PGD extends FGSM by taking iterative steps to produce an adversarial example at the cost of more computation. Again, we generate test cases at five intensity levels with the same max bounds as of FGSM.
\end{itemize}

\begin{figure*}
    \centering
    \includegraphics[width=\linewidth]{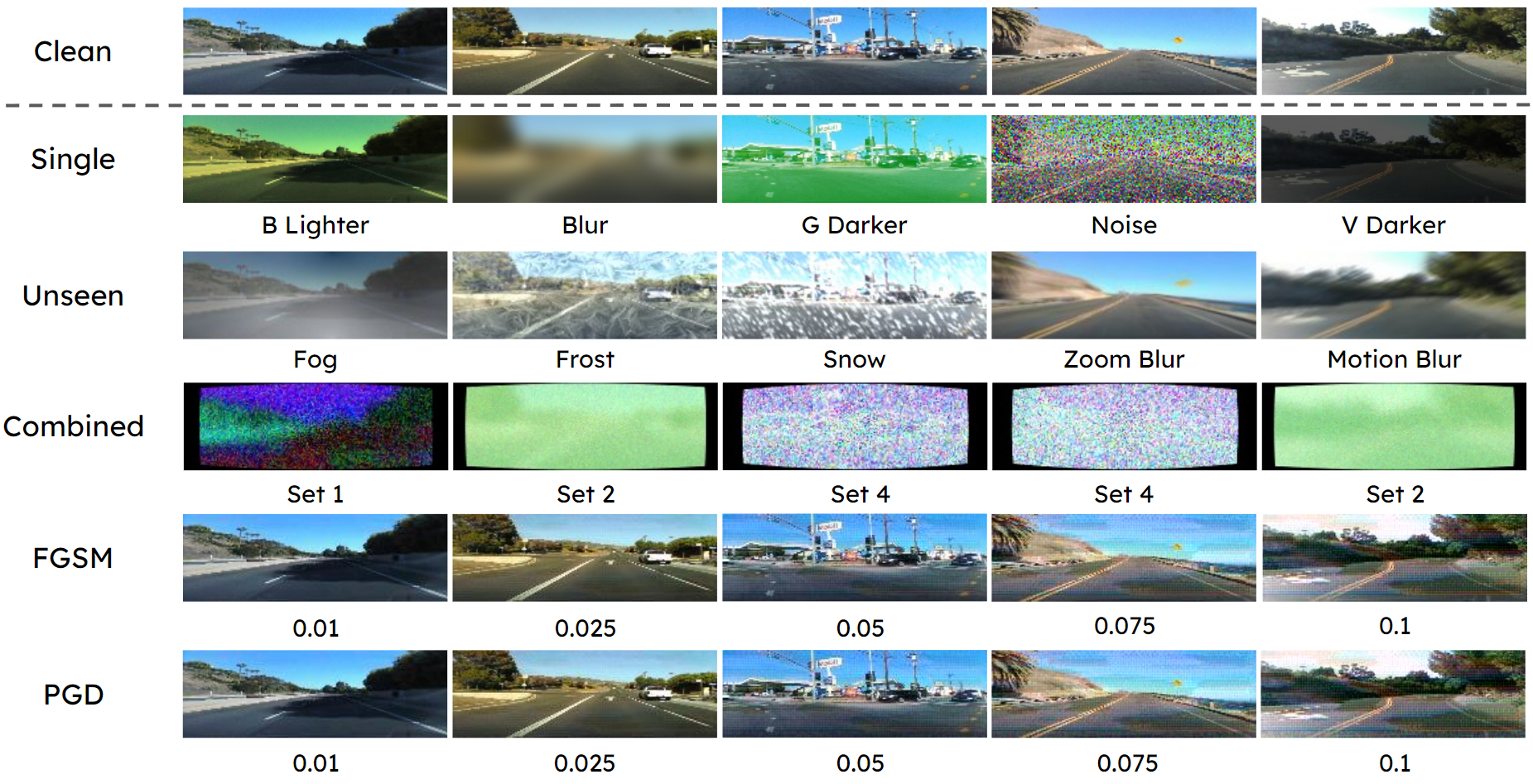}
    \caption{\small{Sample images with perturbations for the six test categories. A column represents a single image that is either clean or perturbed by one of the five perturbation categories. 
    Single images are perturbed by only one of the perturbations outlined in Sec.~\ref{sec:perturbation} Unseen images contain corruptions from ImageNet-C~\cite{hendrycks2019benchmarking}. Combined images have multiple perturbations overlaid, for example, the second column image has G, noise, and blur as the most prominent perturbations. FGSM and PGD adversarial examples are also shown at increasing intensities. The visual differences are not salient due to the preservation of gradient-based adversarial attack potency. 
    }}
    \label{fig:sample_test}
\end{figure*}

Sample images for each test category are given in Fig.~\ref{fig:sample_test}. For Single and Unseen, perturbed images were selected with intensities from 0.5 to 1.0 to highlight the perturbation.

\textbf{Network architectures.} We test on two backbones, the Nvidia model~\cite{bojarski2016end} and ResNet-50~\cite{he2016deep}.
We empirically split the Nvidia model where the encoder is the first seven layers and the regression head is the last two layers; for ResNet-50, the encoder is the first 49 layers and the regression head is the last fully-connected layer. The decoder is a five-layer network with ReLU activations between each layer and a Sigmoid activation for the final layer.

\section{Maximum and Minimum Intensity}
\label{sec:app_minmaxintensity}

We examine the effects of using different ranges of intensities for the perturbations. The original range of intensities for \model{} is $[c_{min}=0,c_{max}=1)$. We perform two sets of experiments: 1) we change $c_{max}$ to be one of \{0.9, 1.1, 1.2, 1.3, 1.4, 1.5\} while leaving $c_{min}=0$; and 2) we change $c_{min}$ to be one of \{0.1, 0.2, 0.3, 0.4, 0.5\} while leaving $c_{max}=1$. For the first experiment set, we change $c_{max}$ to be primarily greater than one to see if learning on more intense perturbations allows for better performance. We also change $c_{max}$ to 0.9 to see if the model does not have to learn on the full range defined by~\cite{shen2021gradient} and still achieves good performance. For the second experiment set, we increase the minimum to see if it is sufficient to learn on images with either no perturbation or a low intensity perturbation to achieve good performance.

\begin{table*}
    \caption{Comparison results on the SullyChen dataset with the Nvidia model using a different range of intensities. The first results set show using a different maximum intensity value, leaving the minimum value at zero. The second results set show using a different minimum value, leaving the maximum value as one. For both sets, the original range of \model{} achieves the best overall performance.}
  \centering
  \begin{tabular}{lcccccccc}
    \toprule
     & \multicolumn{2}{c}{Clean} & \multicolumn{2}{c}{Single} & \multicolumn{2}{c}{Combined} & \multicolumn{2}{c}{Unseen} \\
    \midrule
     & MA (\%) & MAE & MA (\%) & MAE & MA (\%) & MAE & MA (\%) & MAE \\
    \midrule
    Max 0.9 & 88.66 & 3.09 & 84.64 & 4.46 & 67.77 & 10.11 & 81.01 & 5.35 \\
    Max 1.1 & 88.90 & 3.03 & 85.50 & 4.14 & 67.20 & 10.44 & 81.82 & 5.24  \\
    Max 1.2 & 87.77 & 3.29 & 84.47 & 4.32 & \textbf{67.88} & \textbf{9.88} & 80.83 & 5.43  \\
    Max 1.3 & 87.92 & 3.30 & 84.70 & 4.33 & 67.87 & 9.94 & 81.22 & 5.33  \\
    Max 1.4 & 88.07 & 3.24 & 84.95 & 4.29 & 67.44 & 10.15 & 81.16 & 5.37  \\
    Max 1.5 & 87.74 & 3.24 & 84.56 & 4.29 & 65.57 & 10.85 & 80.98 & 5.39  \\
    \midrule
    Min 0.1 & 88.60 & 3.10 & 85.33 & 4.14 & 67.78 & 10.01 & 80.97 & 5.50  \\
    Min 0.2 & 87.14 & 3.46 & 83.95 & 4.54 & 66.57 & 10.49 & 80.05 & 5.74  \\
    Min 0.3 & 87.14 & 3.31 & 84.27 & 4.32 & 66.35 & 10.60 & 80.49 & 5.44  \\
    Min 0.4 & 87.41 & 3.23 & 84.18 & 4.34 & 66.18 & 10.65 & 80.50 & 5.50  \\
    Min 0.5 & 87.56 & 3.33 & 84.20 & 4.41 & 65.58 & 10.87 & 80.14 & 5.62  \\
    \midrule
    \model{} & \textbf{89.46} & \textbf{2.86} & \textbf{86.90} & \textbf{3.53} & 64.67 & 11.21 & \textbf{81.86} & \textbf{5.12} \\
    \bottomrule
  \end{tabular}
\label{ablation_minmax_sully}
\end{table*}

Table~\ref{ablation_minmax_sully} shows the full set of results for SullyChen using the Nvidia architecture. When changing $c_{max}$, the value of 1.1 achieves the most similar performance compared to the original range of \model{}; however, it still performs worse than the original range overall. When looking at changing $c_{min}$, the value of 0.1 results in the closest performance to the original range; however, it also fails to outperform the original range. Looking at both sets of results, changing either $c_{min}$ or $c_{max}$ tends to result in the same magnitude of worse performance for the Clean and Single test categories. However, they differ in that changing $c_{min}$ results in worse performance overall in Unseen for both MA and MAE. These results show a potential vulnerability of the original range as they all outperform the original range in Combined with $c_{max}$ being equal to 1.2 showing the best performance in that category. The results for changing the maximum value show that it is not necessarily the case that learning on more intense perturbations will lead to overall better performance. This could be because the perturbations become intense enough that information necessary for steering angle prediction is lost. The results for changing the minimum value show that it is important for the model to learn on images with no perturbation or a low intensity perturbation given that a minimum of 0.1 achieves the best performance within the set. Overall, however, the original range of \model{} achieves the best prediction performance.

\begin{table*}
    \caption{Comparison results on the A2D2 dataset with the Nvidia model using a different range of intensities. The first results set show using a different maximum intensity value, leaving the minimum value at zero. The second results set show using a different minimum value, leaving the maximum value as one. For both sets, the original range of \model{} achieves the best overall performance.}
  \centering
  \begin{tabular}{lcccccccc}
    \toprule
     & \multicolumn{2}{c}{Clean} & \multicolumn{2}{c}{Single} & \multicolumn{2}{c}{Combined} & \multicolumn{2}{c}{Unseen} \\
    \midrule
     & MA (\%) & MAE & MA (\%) & MAE & MA (\%) & MAE & MA (\%) & MAE \\
    \midrule
    Max 0.9 & 83.82 & 7.16 & 82.13 & 7.65 & 74.06 & 9.79 & 79.04 & 8.79 \\
    Max 1.1 & 83.95 & 7.17 & 82.78 & 7.54 & 78.51 & 8.87 & 79.59 & 8.63  \\
    Max 1.2 & 84.50 & 7.12 & 83.34 & 7.45 & 79.28 & 8.50 & 80.28 & 8.62  \\
    Max 1.3 & \textbf{84.90} & 7.01 & 83.60 & 7.38 & 79.37 & 8.65 & \textbf{80.63} & 8.38  \\
    Max 1.4 & 84.53 & 7.06 & 83.48 & 7.36 & \textbf{79.63} & \textbf{8.41} & 80.59 & 8.26  \\
    Max 1.5 & 84.65 & 6.86 & 83.39 & 7.26 & 79.50 & 8.57 & 80.37 & 8.30  \\
    \midrule
    Min 0.1 & 83.74 & 7.04 & 82.22 & 7.61 & 73.89 & 10.84 & 79.32 & 8.71  \\
    Min 0.2 & 84.29 & 7.17 & 83.19 & 7.50 & 77.70 & 9.28 & 79.91 & 8.69  \\
    Min 0.3 & 84.16 & 7.35 & 83.12 & 7.61 & 77.16 & 9.16 & 79.77 & 8.82  \\
    Min 0.4 & 83.80 & 7.33 & 82.82 & 7.60 & 77.00 & 9.20 & 79.17 & 8.98  \\
    Min 0.5 & 84.06 & 7.41 & 82.93 & 7.72 & 75.89 & 10.17 & 79.33 & 9.09  \\
    \midrule
    \model{} & 84.70 & \textbf{6.79} & \textbf{83.70} & \textbf{7.07} & 79.12 & 8.58 & 80.31 & \textbf{8.23} \\
    \bottomrule
  \end{tabular}
\label{ablation_minmax_audi}
\end{table*}

The results for A2D2, shown in Table~\ref{ablation_minmax_audi}, are more inconsistent than SullyChen given that the original range is outperformed in four columns instead of just two. The original range is outperformed by changing $c_{max}$ to 1.3 for the Clean MA and Unseen MA columns and changing $c_{max}$ to 1.4 for Combined. Table~\ref{ablation_minmax_audi} does show, similar to Table~\ref{ablation_minmax_sully}, that learning on more intense perturbations will necessarily lead to better performance when the test intensities are left unchanged. These results also show it is important to learn on images without a perturbation/low intensity perturbation given that the original range outperforms all of the experiments when changing the minimum value. When examining both Table~\ref{ablation_minmax_sully} and Table~\ref{ablation_minmax_audi}, the only times the new ranges outperform the original range of \model{} is when $c_{max}$ is increased. However, for both SullyChen and A2D2, the original range achieves the best overall performance.

\section{Feedback Loop}
\label{sec:app_feedback_loop}

This section contains results and discussion for the SullyChen, A2D2, Honda, and Waymo datasets. Adding the denoised images results in adding a third term to Eq.~\ref{loss_function}:  

\begin{equation}
    \mathcal{L} = \lambda_1\ell_2 \left(\mathbf{x'_i},\mathbf{x_i}\right)+\lambda_2 \ell_1 \left(\mathbf{a_{p_{i}}},\mathbf{a_{t_{i}}}\right)+\lambda_3 \ell_1 
    (\mathbf{a_{p'_{i}}},\mathbf{a_{t_{i}}}),
\label{loss_function_feedback}
\end{equation}

where $\lambda_3$ is the weight of the new term and $a_{p'_{i}}$ is the predicted steering angle on the reconstruction $x'_{i}$. 

\begin{table*}
    \caption{\small{Results on the SullyChen dataset with the Nvidia backbone and including the feedback loop. The weight coefficients are presented in the order of terms of Eq.~\ref{loss_function_feedback}. 
    Given the overall decreased performance, we exclude the feedback loop from \model{}.}}
  \centering
  \addtolength{\tabcolsep}{-0.2em}

  \begin{tabular}{lcccccccc}
    \toprule
     & \multicolumn{2}{c}{Clean} & \multicolumn{2}{c}{Single} & \multicolumn{2}{c}{Combined} & \multicolumn{2}{c}{Unseen} \\
    \midrule
     & MA (\%) & MAE & MA (\%) & MAE & MA (\%) & MAE & MA (\%) & MAE \\
    \midrule
    (10,1,1) & 86.93 & 3.56 & 83.60 & 4.66 & 64.10 & 11.10 & 78.88 & 6.25 \\
    (1,10,1) & 89.11 & 3.07 & 85.60 & 4.15 & \textbf{68.23} & \textbf{9.70} & 81.58 & 5.27  \\
    (1,1,10) & 81.34 & 5.08 & 78.77 & 6.04 & 64.36 & 11.21 & 76.31 & 6.71  \\
    \midrule
    \model{} (1,10) & \textbf{89.46} & \textbf{2.86} & \textbf{86.90} & \textbf{3.53} & 64.67 & 11.21 & \textbf{81.86} & \textbf{5.12} \\
    \bottomrule
  \end{tabular}
\label{feedback_sully}
\end{table*} 

Looking at Table~\ref{feedback_sully}, emphasizing the reconstruction regression loss causes significant performance loss compared to \model{} with 8.12\% MA/2.22 MAE and 8.13\% MA/2.51 MAE decreases on Clean and Single, respectively. 
This suggests the data contained within the reconstructions is detrimental to the overall performance/robust capabilities of the regression model. Emphasizing reconstruction loss results in worse performance than \model{}, which is expected as \model{} emphasizes regression loss for the SullyChen dataset. 
Emphasizing the regression loss results in improvements in Combined (3.56\% MA and 1.51 MAE) at the cost of detriment to performance in all other categories. Overall, there is a decrease in performance when adding the feedback loop.

\begin{table*}
    \caption{Comparison results on the A2D2 dataset with the Nvidia model using different subsets of the original set of perturbations. The weight coefficients are presented in the order: reconstruction loss, regression loss, reconstruction regression loss. \model{}'s original set of weights outperforms all three weight coefficient tuples.}
  \centering
  \addtolength{\tabcolsep}{-0.2em}
  \begin{tabular}{lcccccccc}
    \toprule
     & \multicolumn{2}{c}{Clean} & \multicolumn{2}{c}{Single} & \multicolumn{2}{c}{Combined} & \multicolumn{2}{c}{Unseen} \\
    \midrule
     & MA (\%) & MAE & MA (\%) & MAE & MA (\%) & MAE & MA (\%) & MAE \\
    \midrule
    (10,1,1) & 83.98 & 7.07 & 82.81 & 7.44 & 78.63 & 8.75 & 79.26 & 8.95 \\
    (1,10,1) & 84.18 & 7.05 & 83.09 & 7.38 & 77.80 & 8.81 & 79.72 & 8.57  \\
    (1,1,10) & 83.01 & 7.42 & 81.89 & 7.78 & 77.21 & 8.97 & 79.67 & 8.39  \\
    \midrule
    \model{} (1,10) & \textbf{84.70} & \textbf{6.79} & \textbf{83.70} & \textbf{7.07} & \textbf{79.12} & \textbf{8.58} & \textbf{80.31} & \textbf{8.23} \\
    \bottomrule
  \end{tabular}
\label{ablation_feedback_audi}
\end{table*}

In Table~\ref{ablation_feedback_audi}, a similar trend to Table~\ref{feedback_sully} is seen where the weight tuple (1,10,1) is the best performing of the three weight tuples, while the tuple (1,1,10) offers the worst performance. This reiterates that emphasizing the reconstruction regressions had detrimental effects, which is potentially due to information loss within the reconstructions. However, A2D2 is less affected by adding an additional loss term compared to SullyChen. This is seen as the differences in ranges of performance between the weight coefficients are much greater for SullyChen than for A2D2. For example, the range of performance on Clean for SullyChen is 7.77\% MA/2.01 MAE while for A2D2, it is just 1.17\% MA/0.37 MAE. Overall, the original weights of \model{} provide for the best performance.

\begin{table*}
    \caption{Comparison results on the Honda dataset with the ResNet-50 model using different subsets of the original set of perturbations. The weight coefficients are presented in the order: reconstruction loss, regression loss, reconstruction regression loss. Adding the feedback loop for the Honda dataset, results in significant performance loss for all three weight tuples. Because of this, the original weights of \model{} achieve the best performance.}
  \centering
  \addtolength{\tabcolsep}{-0.2em}
  \begin{tabular}{lcccccccc}
    \toprule
     & \multicolumn{2}{c}{Clean} & \multicolumn{2}{c}{Single} & \multicolumn{2}{c}{Combined} & \multicolumn{2}{c}{Unseen} \\
    \midrule
     & MA (\%) & MAE & MA (\%) & MAE & MA (\%) & MAE & MA (\%) & MAE \\
    \midrule
    (10,1,1) & 79.58 & 6.53 & 77.19 & 8.13 & 65.19 & 18.21 & 77.15 & 8.84 \\
    (1,10,1) & 85.70 & 3.53 & 83.23 & 4.67 & 61.41 & 20.68 & 81.13 & 5.72  \\
    (1,1,10) & 51.30 & 14.98 & 49.19 & 16.79 & 42.71 & 22.15 & 49.15 & 17.15  \\
    \midrule
    \model{} (1,10) & \textbf{96.46} & \textbf{1.12} & \textbf{94.58} & \textbf{1.98} & \textbf{70.70} & \textbf{14.56} & \textbf{91.92} & \textbf{2.89} \\
    \bottomrule
  \end{tabular}
\label{ablation_feedback_honda}
\end{table*}

Table~\ref{ablation_feedback_honda}) shows the results for Honda with ResNet-50. The results show that adding the feedback loop for Honda results in significant performance loss, even when emphasizing regression loss. For example, the greatest differences between the three experiment weight tuples and \model{}'s original weight tuple are 8.12\% MA and 2.22 MAE for SullyChen and 1.91\% MA and 0.71 MAE for A2D2. However, the least differences between the three weight tuples and \model{}'s original weight tuple for Honda is 9.29\% MA and 2.41 MAE. Emphasizing the reconstruction regression loss results in significant performance losses of 45.16\% MA and 13.86 MAE in Clean, which is a 47\% decrease for MA and a 93\% decrease for MAE; Single, Combined, and Unseen also have significant performance losses. Emphasizing regression loss also results in significant performance loss such as 11.35\% MA and 2.69 MAE decreases in Single; these equates to a 14\% MA decrease and 57\% MAE decrease. These significant performance losses are part of our reasoning to exclude the feedback loop as a main component of \model{}. \model{} performs better on Honda without the feedback loop.


\begin{table*}
    \caption{Comparison results on the Waymo dataset with the ResNet-50 model using different subsets of the original set of perturbations. The weight coefficients are presented in the order: reconstruction loss, regression loss, reconstruction regression loss. Emphasizing the reconstruction regression loss term results in SOTA performance.}
  \centering
  \addtolength{\tabcolsep}{-0.2em}
  \begin{tabular}{lcccccccc}
    \toprule
     & \multicolumn{2}{c}{Clean} & \multicolumn{2}{c}{Single} & \multicolumn{2}{c}{Combined} & \multicolumn{2}{c}{Unseen} \\
    \midrule
     & MA (\%) & MAE & MA (\%) & MAE & MA (\%) & MAE & MA (\%) & MAE \\
    \midrule
    (10,1,1) & 63.14 & 19.15 & 63.38 & 19.11 & 57.98 & 23.15 & 61.61 & 20.41 \\
    (1,10,1) & 63.35 & 18.89 & 63.22 & 19.37 & 56.32 & 35.88 & 62.32 & 21.31  \\
    (1,1,10) & \textbf{67.70} & 18.00 & \textbf{66.68} & \textbf{18.28} & \textbf{67.70} & \textbf{18.00} & \textbf{67.70} & \textbf{18.00}  \\
    \midrule
    \model{} (10,1) & 64.91 & 18.02 & 63.84 & 19.30 & 58.74 & 26.42 & 64.17 & 19.10 \\
    Fuse (10,1) & 65.07 & \textbf{17.60} & 64.34 & 18.49 & 63.48 & 20.82 & 65.01 & 18.17  \\
    \bottomrule
  \end{tabular}
\label{ablation_feedback_waymo}
\end{table*}

Table~\ref{ablation_feedback_waymo} shows the results for adding the feedback loop to Waymo on ResNet-50. Waymo uses a weight tuple of (10,1) in \model{} for better performance, while the other datasets use (1,10). This suggests that learning on the underlying distribution of the data and the reconstructions provide significant benefit over emphasizing regression loss. However, when adding the feedback loop, emphasizing the regression loss results in better performance than emphasizing the reconstruction loss; however, both are outperformed by \model{}. The performance trend for Waymo is significantly different from the other datasets as emphasizing the reconstruction regression loss results in SOTA performance. \model{}-Fuse's results are shown for further comparison since it is the SOTA within the main text. This is an intriguing development because of the negative performance impacts that the feedback loop has on the other datasets. This result is further evidence towards the idea the learning underlying distributions of Waymo leads to better performance. Overall, emphasizing the reconstruction regression results in SOTA performance.

\section{Full Set of Perturbations Not Guaranteed and No Random Intensities}
\label{sec:app_nine_static}

From Table~\ref{results_no_dae_sully}, \model{} without the denoising autoencoder (DAE) already outperforms the work by~\cite{shen2021gradient}. Outside of adding the DAE, the main changes from their work to our work is that we ensure that all 15 perturbations are seen during learning and that the intensities are sampled from a range instead of using distinct intensities. Thus, we want to examine if these changes have an effect on performance and can account for the reason that \model{} without the DAE outperforms the Shen model. We break down this set of experiments into three sets of cases: 1) not guaranteeing the full set of 15 perturbations are seen by the model, 2) not using random intensities, or 3) both. The original methodology of \model{} is left the same except for the changes of each case. The third case brings the methodology of \model{} closest to that of the work by~\cite{shen2021gradient} although they are not the same entirely. 

The first case is accomplished by not discretizing the single channel perturbations as described in Sec.~\ref{sec:perturbation}. Whether to lighten or darken the R, G, B, H, S, and V channels of the images is decided stochastically. This means there is potential the model does not see all 15 perturbations, although highly unlikely; however, it is highly likely that the model does not see them with the same frequency as with the original methodology of \model{}. The second case is done by using the five distinct intensities from the work by~\cite{shen2021gradient}, which are \{0.02, 0.2, 0.5, 0.65, 1.0\}. The intensity for a perturbation is still sampled from within this set of values, but it is inherently not as wide of a distribution space compared to the methodology described in Sec.~\ref{sec:perturbation}. The third case combines the changes in procedure outlined above.

\begin{table*}
    \caption{Comparison results on the SullyChen dataset with the Nvidia model looking at the cases where it is not guaranteed the \textbf{F}ull \textbf{S}et of perturbations is seen by the model, not using \textbf{R}andom \textbf{I}ntensities, or both. The distinct intensities come from Shen. Using the original \model{} setup results in the best overall performance across all subsets of perturbations.}
  \centering
  \addtolength{\tabcolsep}{-0.2em}
  \begin{tabular}{lcccccccc}
    \toprule
     & \multicolumn{2}{c}{Clean} & \multicolumn{2}{c}{Single} & \multicolumn{2}{c}{Combined} & \multicolumn{2}{c}{Unseen} \\
    \midrule
     & MA (\%) & MAE & MA (\%) & MAE & MA (\%) & MAE & MA (\%) & MAE \\
    \midrule
    w/o FS & 87.74 & 3.32 & 84.54 & 4.34 & 66.27 & 10.32 & 80.62 & 5.50 \\
    w/o RI & 88.07 & 3.42 & 84.76 & 4.42 & \textbf{67.72} & \textbf{10.12} & 80.92 & 5.65  \\
    w/o FS+RI & 86.43 & 3.54 & 83.19 & 4.62 & 61.97 & 13.01 & 78.51 & 6.23  \\
    \midrule
    \model{} & \textbf{89.46} & \textbf{2.86} & \textbf{86.90} & \textbf{3.53} & 64.67 & 11.21 & \textbf{81.86} & \textbf{5.12} \\
    \bottomrule
  \end{tabular}
\label{ablation_ninestatic_sully}
\end{table*}

\begin{table*}
    \caption{Comparison results on the A2D2 dataset with the Nvidia model looking at the cases where it is not guaranteed the \textbf{F}ull \textbf{S}et of perturbations is seen by the model, not using \textbf{R}andom \textbf{I}ntensities, or both. Using the original \model{} setup results in the best overall performance across all subsets of perturbations.}
  \centering
  \addtolength{\tabcolsep}{-0.2em}
  \begin{tabular}{lcccccccc}
    \toprule
     & \multicolumn{2}{c}{Clean} & \multicolumn{2}{c}{Single} & \multicolumn{2}{c}{Combined} & \multicolumn{2}{c}{Unseen} \\
    \midrule
     & MA (\%) & MAE & MA (\%) & MAE & MA (\%) & MAE & MA (\%) & MAE \\
    \midrule
    w/o FS & 83.93 & 7.24 & 82.77 & 7.52 & 78.60 & 8.90 & 78.45 & 9.78 \\
    w/o RI & 83.90 & 6.95 & 82.68 & 7.35 & 78.20 & 8.72 & 78.38 & 9.20  \\
    w/o FS+RI & 83.90 & 7.10 & 82.85 & 7.42 & 78.45 & 8.63 & 79.01 & 9.05  \\
    \midrule
    \model{} & \textbf{84.70} & \textbf{6.79} & \textbf{83.70} & \textbf{7.07} & \textbf{79.12} & \textbf{8.58} & \textbf{80.31} & \textbf{8.23} \\
    \bottomrule
  \end{tabular}
\label{ablation_ninestatic_audi}
\end{table*}

Looking at the effects the three cases have on SullyChen and A2D2 using the Nvidia model, the results show that ensuring all 15 perturbations are seen during learning and sampling the intensities does improve overall performance when predicting steering angles and these changes are significant to the training of the model. This gives more credence to why \model{} without the DAE is able to outperform Shen. 

The effects of the three cases differs between the two datasets. Table~\ref{ablation_ninestatic_sully} shows that using both is able to significantly impact performance on all categories by decreasing performance by an average of 3.20\% MA and 1.17 MAE across all test categories for SullyChen. Using distinct intensities allows for significantly better performance on Combined (the model without FS also achieves better performance in this category), but fails to outperform in Clean, Single, and Unseen categories. For A2D2, the overall effect is much less severe as the differences between the three effects and \model{}'s methodology are in closer proximity with roughly a difference of 1.0\% MA and 0.5 MAE. However, the original setup still results in the overall best steering angle prediction performance.

\begin{table*}
    \caption{Comparison results on the Honda dataset with the ResNet-50 model looking at the cases where it is not guaranteed the \textbf{F}ull \textbf{S}et of perturbations is seen by the model, not using \textbf{R}andom \textbf{I}ntensities, or both. The model without FS results in the best overall performance of the model, which is different from the SullyChen, A2D2, and Waymo datasets.}
  \centering
  \addtolength{\tabcolsep}{-0.2em}
  \begin{tabular}{lcccccccc}
    \toprule
     & \multicolumn{2}{c}{Clean} & \multicolumn{2}{c}{Single} & \multicolumn{2}{c}{Combined} & \multicolumn{2}{c}{Unseen} \\
    \midrule
     & MA (\%) & MAE & MA (\%) & MAE & MA (\%) & MAE & MA (\%) & MAE \\
    \midrule
    w/o FS & \textbf{96.78} & \textbf{1.05} & \textbf{95.17} & \textbf{1.82} & \textbf{75.16} & \textbf{12.34} & 91.69 & 3.27 \\
    w/o RI & 96.53 & 1.08 & 94.92 & 1.86 & 68.63 & 17.14 & 91.53 & 3.18  \\
    w/o FS+RI & 96.72 & 1.06 & 95.20 & 1.80 & 65.49 & 24.44 & 90.97 & 3.91  \\
    \midrule
    \model{} & 96.46 & 1.12 & 94.58 & 1.98 & 70.70 & 14.56 & \textbf{91.92} & \textbf{2.89} \\
    \bottomrule
  \end{tabular}
\label{ablation_ninestatic_honda}
\end{table*}

Table~\ref{ablation_ninestatic_honda} shows the results for Honda with ResNet-50. Unlike SullyChen and A2D2, all three cases actually outperform \model{} for both Clean and Single. \model{} is even outperformed in Combined when not ensuring the full set. Not ensuring the full set has potential for more variability of when perturbations are learned by the model, which can increase the perturbation distribution space allowing for better generalization. However, when not ensuring the full set and using distinct intensities, there is a loss of generalization as \model{} outperforms this case in Combined and Unseen. The Shen model outperforms \model{} in Clean and Combined MAE. Shen still outperforms the case of not ensuring the full set on Clean, but the Shen model is outperformed on Combined MAE. \model{} achieves the best performance in Unseen amongst all the three cases; however, overall the model without the full set provides for the best performance.

\begin{table*}
    \caption{Comparison results on the Waymo dataset with the ResNet-50 model looking at the cases where it is not guaranteed the \textbf{F}ull \textbf{S}et of perturbations is seen by the model, not using \textbf{R}andom \textbf{I}ntensities, or both. Using all of them results in the best overall performance across all subsets of perturbations.}
  \centering
  \addtolength{\tabcolsep}{-0.2em}
  \begin{tabular}{lcccccccc}
    \toprule
     & \multicolumn{2}{c}{Clean} & \multicolumn{2}{c}{Single} & \multicolumn{2}{c}{Combined} & \multicolumn{2}{c}{Unseen} \\
    \midrule
     & MA (\%) & MAE & MA (\%) & MAE & MA (\%) & MAE & MA (\%) & MAE \\
    \midrule
    W/o FS & 63.95 & 18.40 & 63.40 & 18.85 & \textbf{61.04} & \textbf{21.46} & 63.40 & 19.15 \\
    W/o RI & 64.12 & 18.57 & 63.76 & 19.14 & 54.80 & 32.87 & 62.27 & 21.52  \\
    W/o FS+RI & 63.96 & 18.51 & 63.70 & \textbf{18.76} & 56.25 & 26.81 & 62.79 & 19.81  \\
    \midrule
    \model{} & \textbf{64.91} & \textbf{18.02} & \textbf{63.84} & 19.30 & 58.74 & 26.42 & \textbf{64.17} & \textbf{19.10} \\
    \bottomrule
  \end{tabular}
\label{ablation_ninestatic_waymo}
\end{table*}

The idea that ensuring all 15 perturbations are seen during learning and sampling the perturbation intensities does improve the overall performance of the model returns with Waymo on ResNet-50. Table~\ref{ablation_ninestatic_waymo} shows these results. \model{} outperforms all three cases in Clean, Single MA, and Unseen. It is outperformed by the all three cases in Single MAE and is outperformed by the model without FS in Combined; however, it outperforms the other two cases in Combined. Looking at the results for Honda and Waymo, it appears that not ensuring all 15 perturbations are seen during training provides for the best performance for Combined; however still fails to outperform \model{} in Unseen. These two categories are different from Clean and Single as the model never learns on them during training. The model without FS is able to generalize better for Combined than Unseen given the results.

\section{Perturbation Study}
\label{sec:app_perturb_study}

This section contains more results and discussion for the other datasets of A2D2, Honda, and Waymo for the experiments where different subsets of perturbations are used.

\begin{table*}
    \caption{Comparison results on the A2D2 dataset with the Nvidia model using different subsets of the original set of perturbations. ``No BND'' means that blur, noise, and distort are not used within the perturbation set. The single perturbation column is removed for a fair comparison. Using all of them results in the best overall performance across all subsets of perturbations.}
  \centering
  \begin{tabular}{lcccccc}
    \toprule
     & \multicolumn{2}{c}{Clean} & \multicolumn{2}{c}{Combined} & \multicolumn{2}{c}{Unseen} \\
    \midrule
     & MA (\%) & MAE & MA (\%) & MAE & MA (\%) & MAE \\
    \midrule
    No RGB & 84.46 & 6.61 & 77.08 & 9.51 & 79.67 & 8.65 \\
    No HSV & 84.50 & 6.70 & 76.51 & 9.41 & 78.01 & 9.36  \\
    No BND & 83.72 & 7.21 & 68.88 & 12.09 & 78.49 & 9.13  \\
    RGB & 83.82 & 7.20 & 67.49 & 12.40 & 76.93 & 9.87 \\
    HSV & 83.19 & 7.26 & 67.91 & 12.65 & 78.00 & 9.31 \\
    Only RGB+Noise & 83.92 & 6.91 & 73.56 & 10.03 & 78.44 & 8.96 \\
    Only HSV+Noise & 84.39 & 6.87 & 70.96 & 11.47 & 79.53 & 8.61  \\
    No Blur,Distort & 82.53 & 7.49 & 74.99 & 9.53 & 77.40 & 9.57 \\
    \midrule
    All & \textbf{84.70} & \textbf{6.79} & \textbf{79.12} & \textbf{8.58} & \textbf{80.31} & \textbf{8.23} \\
    \bottomrule
  \end{tabular}
\label{ablation_perturb_audi}
\end{table*}

The trends for A2D2 using the Nvidia model are different compared to SullyChen. The results are given in Table~\ref{ablation_perturb_audi}. While A2D2 is similar to SullyChen in that the best performance comes from using all of the perturbations, the model with no BND perturbations performs the worst on Combined implying that some combination of these perturbations is important for model generalizability for Combined. This idea is aided by the other scenarios where performance on Combined is improved when Gaussian noise is added back to the set of perturbations seen by the model. The closest in overall performance to using all perturbations is not using RGB perturbations within the training set. For Unseen, there are no clear patterns within the performances amongst the various subsets with the worst performing subset being not using blur and distort perturbations at 77.40\% MA and 9.57 MAE. The other trend that is similar to SullyChen, however, is that Combined contains the most volatility in the performance; the range from the worst performing subset to the best performing subset is 68.88\% MA and 12.09 MAE to 79.12\% MA and 8.58 MAE. Using all perturbations during learning results in the best performance.

\begin{table*}
    \caption{Comparison results on the Honda dataset with the ResNet-50 model using different subsets of the original set of perturbations. ``No BND'' means that blur, noise, and distort are not used within the perturbation set. The single perturbation column is removed for a fair comparison. Using all perturbations is overall the best performing model despite being outperformed in the Clean and Combined categories.}
  \centering
  \begin{tabular}{lcccccc}
    \toprule
     & \multicolumn{2}{c}{Clean} & \multicolumn{2}{c}{Combined} & \multicolumn{2}{c}{Unseen} \\
    \midrule
     & MA (\%) & MAE & MA (\%) & MAE & MA (\%) & MAE \\
    \midrule
    No RGB & \textbf{96.78} & \textbf{1.02} & 66.37 & 21.67 & 91.72 & 3.26 \\
    No HSV & 96.48 & 1.07 & \textbf{74.96} & \textbf{13.19} & 88.25 & 5.67  \\
    No BND & 96.08 & 1.27 & 63.88 & 18.97 & 90.58 & 3.55  \\
    RGB & 95.75 & 1.38 & 44.90 & 32.36 & 83.53 & 7.32 \\
    HSV & 95.94 & 1.31 & 53.99 & 29.83 & 83.53 & 7.59 \\
    Only RGB+Noise & 96.39 & 1.13 & 69.21 & 14.48 & 87.13 & 5.70 \\
    Only HSV+Noise & 96.47 & 1.12 & 69.08 & 14.72 & 91.77 & 2.93  \\
    No Blur,Distort & 96.33 & 1.17 & 67.74 & 14.73 & 91.16 & 3.15 \\
    \midrule
    All & 96.46 & 1.12 & 70.70 & 14.56 & \textbf{91.92} & \textbf{2.89} \\
    \bottomrule
  \end{tabular}
\label{ablation_perturb_honda}
\end{table*}

Table~\ref{ablation_perturb_honda} shows the results for Honda on ResNet-50. Using all perturbations is outperformed several cases in Clean and Combined. Not using RGB perturbations achieves the best performance in Clean and not using HSV perturbations achieves the best performance in Combined (by a significant margin of 4.26\% MA/1.37 MAE). Even with the clean performance increases, the Shen model is still the best in Clean. Well-defined patterns are still not clear in the results. Not using RGB perturbations performs worse than using all perturbations in Combined. Not using HSV perturbations significantly improves performance in Combined at a 4.26\% MA and 1.37 MAE improvement; however, results in a significant decrease in performance in Unseen with 3.67\% MA and 2.78 MAE detriments. The range of performance for Combined is the largest compared to the other categories. This is similar to SullyChen and A2D2, showing further evidence of the volatility within Combined. The closest in performance to using all perturbations is not using HSV perturbations; this case results in a net gain of 0.61\% MA and a net loss of 1.36 MAE when compared to using all. Given that MAE, for Honda, are small values and lie within a tighter range than MA, the net loss of 1.36 MAE means that using all perturbations is actually the best performing model overall.

\begin{table*}
    \caption{Comparison results on the Waymo dataset with the ResNet-50 model using different subsets of the original set of perturbations. ``No BND'' means that blur, noise, and distort are not used within the perturbation set. The single perturbation column is removed for a fair comparison. Using all of them results in the best overall performance across all subsets of perturbations.}
  \centering
  \begin{tabular}{lcccccc}
    \toprule
     & \multicolumn{2}{c}{Clean} & \multicolumn{2}{c}{Combined} & \multicolumn{2}{c}{Unseen} \\
    \midrule
     & MA (\%) & MAE & MA (\%) & MAE & MA (\%) & MAE \\
    \midrule
    No RGB & 64.63 & 18.20 & 60.38 & 24.79 & 63.63 & 19.72 \\
    No HSV & 63.94 & 18.46 & \textbf{61.18} & \textbf{20.89} & 63.09 & 20.02  \\
    No BND & 64.56 & 18.06 & 50.21 & 43.28 & 63.06 & 20.13  \\
    RGB & 64.64 & 18.12 & 49.32 & 36.31 & 62.27 & 20.25 \\
    HSV & \textbf{65.00} & 18.37 & 52.03 & 34.21 & 63.85 & 19.52 \\
    Only RGB+Noise & 63.95 & 18.40 & 52.84 & 31.01 & 60.70 & 23.43 \\
    Only HSV+Noise & 64.48 & 17.97 & 59.97 & 24.39 & 63.65 & 19.37  \\
    No Blur,Distort & 64.04 & 18.06 & 57.29 & 28.48 & 62.89 & 19.91 \\
    \midrule
    All & 64.91 & \textbf{18.02} & 58.74 & 26.42 & \textbf{64.17} & \textbf{19.10} \\
    \bottomrule
  \end{tabular}
\label{ablation_perturb_waymo}
\end{table*}

Table~\ref{ablation_perturb_waymo} shows the results for Waymo with ResNet-50. Using all perturbations results in the best overall performance for the model, although not using HSV perturbations outperforms using all in Combined for both MA and MAE. Combined has the widest range in performances amongst the subsets confirming that Combined, in general, is the most volatile in performance across all the datasets used. Not using RGB perturbations, not using HSV perturbations, and only using HSV perturbations and Gaussian noise outperform using all perturbations in Combined; however, this does not translate over to Clean and Unseen. Only using RGB and Gaussian noise perturbations results in the overall worst performance across the three categories, but any further patterns can not be well-defined from this as using RGB perturbations and/or Gaussian noise in other cases results in relatively good performance. Overall, using all perturbations results in the best performing prediction model.

\section{Times For Experiments}
\label{sec:app_time_tables}

We present Tables~\ref{ablation_time_rn50} and~\ref{ablation_time_nvidia} with times for various experiments. Table~\ref{ablation_time_rn50} shows time (in seconds) per epoch for Standard, AugMix, Shen, and AutoJoin The experiments are on AutoJoin, Shen, AugMix, and Standard with both the ResNet-50 and Nvidia models. Standard is given to show baseline efficiency when not performing any robustness training. From the table, \model{} is the most efficient compared to the other techniques. AugMix is close in efficiency as it is within at most 10 seconds of our technique's time. Shen's efficiency is significantly worse compared to both \model{} and AugMix as it is many times slower than both techniques. Note that for both tables, Shen's time does not reflect a selection process that occurs in-between training that results in additional training time.

\begin{table}
    \caption{Table comparing the efficiency of different techniques in terms of time (in seconds) per each epoch on the ResNet-50 model. \model{} is the most efficient out of all the techniques.}
  \centering
  \begin{tabular}{lcc}
    \toprule
     & Honda & Waymo \\
    \midrule
    Standard & 90 & 97  \\
    \midrule
    AugMix & 118 & 128  \\
    Shen & 759 & 818  \\
    \model{} & \textbf{109} & \textbf{118}  \\
    \bottomrule
  \end{tabular}
\label{ablation_time_rn50}
\end{table}

Table~\ref{ablation_time_nvidia} shows the times for Standard, AugMix, Shen, and \model{} using the Nvidia model. Standard is given to provide a baseline efficiency when not performing any robustness training. \model{} still achieves the best efficiency out of the techniques; however, Shen is more efficient than AugMix, which is not the case in Table~\ref{ablation_time_rn50}. This suggests that with smaller datasets, Shen's technique is able to maintain efficiency and as the datasets starts growing, Shen's efficiency significantly decreases. 

\begin{table}
    \caption{Table comparing the efficiency of different techniques in terms of time (in seconds) per each epoch on the Nvidia model. \model{} is the most efficient out of all the techniques.}
  \centering
  \begin{tabular}{lcc}
    \toprule
     & SullyChen & A2D2 \\
    \midrule
    Standard & 2 & 4  \\
    \midrule
    AugMix & 10 & 22  \\
    Shen & 9 & 16  \\
    \model{} & \textbf{5} & \textbf{9}  \\
    \bottomrule
  \end{tabular}
\label{ablation_time_nvidia}
\end{table}

\section{Gradient-based Adversarial Transferability}
\label{app_gradient-based}

\begin{table*}
  \caption{\small{Results on gradient-based adversarial examples using the A2D2 dataset and the Nvidia backbone. Each column represents a dataset generated at a certain intensity of FGSM/PGD (higher values mean higher intensities). All results are in MA (\%). \model{} achieves the least adversarial transferability among all techniques tested under all intensities of FGSM~\cite{goodfellow2014explaining} and PGD~\cite{madry2017towards}.}}
  \centering
  \addtolength{\tabcolsep}{-0.2em}
  \begin{tabular}{lccccc}
    \toprule 
     & \multicolumn{5}{c}{FGSM} \\
    \midrule
     & 0.01 & 0.025 & 0.05 & 0.075 & 0.1 \\
    \midrule
    Standard & 73.91 & 65.42 & 57.70 & 53.27 & 50.12  \\
    AdvBN & 76.34 & 76.14 & 75.50 & 74.25 & 72.75  \\
    AugMix & 77.66 & 76.69 & 73.61 & 69.74 & 66.38  \\
    AugMax & 77.04 & 76.94 & 76.18 & 75.10 & 73.91  \\
    MaxUp & 78.71 & 78.47 & 78.10 & 77.42 & 76.71  \\
    Shen & 80.10 & 79.83 & 79.02 & 77.94 & 76.98 \\
    \midrule
    \model{} & \textbf{84.11} & \textbf{83.83} & \textbf{83.13} & \textbf{82.02} & \textbf{81.14} \\
    \bottomrule
  \end{tabular}
  \begin{tabular}{ccccc}
    \toprule
      \multicolumn{5}{c}{PGD} \\
    \midrule
    0.01 & 0.025 & 0.05 & 0.075 & 0.1 \\
    \midrule
    73.87 & 65.60 & 57.93 & 53.43 & 51.07 \\
    76.35 & 76.17 & 75.62 & 74.46 & 72.91 \\
    77.65 & 76.75 & 73.74 & 69.75 & 66.40  \\
    77.04 & 76.93 & 76.23 & 75.10 & 73.91 \\
    78.71 & 78.47 & 78.09 & 77.39 & 76.72  \\
    80.09 & 79.79 & 79.02 & 77.93 & 76.94 \\
    \midrule
    \textbf{84.14} & \textbf{83.84} & \textbf{83.15} & \textbf{81.97} & \textbf{81.09}\\
    \bottomrule
  \end{tabular}
\label{results_transfer_audi_nvidia}
\end{table*}

Although \model{} is a gradient-free technique with the focus on gradient-free attacks, we curiously test it on gradient-based adversarial examples. Dataset details and sample images are given in Appendix~\ref{app:exp}
The evaluation results using the A2D2 dataset and the Nvidia backbone are shown in Table~\ref{results_transfer_audi_nvidia}. 
\model{} surprisingly demonstrates superb ability in defending adversarial transferability against gradient-based attacks by outperforming every other approaches by large margins at all intensity levels of FGSM and PGD.

\end{appendices} 

%% file: main.bbl
\begin{thebibliography}{10}
\providecommand{\url}[1]{#1}
\csname url@samestyle\endcsname
\providecommand{\newblock}{\relax}
\providecommand{\bibinfo}[2]{#2}
\providecommand{\BIBentrySTDinterwordspacing}{\spaceskip=0pt\relax}
\providecommand{\BIBentryALTinterwordstretchfactor}{4}
\providecommand{\BIBentryALTinterwordspacing}{\spaceskip=\fontdimen2\font plus
\BIBentryALTinterwordstretchfactor\fontdimen3\font minus \fontdimen4\font\relax}
\providecommand{\BIBforeignlanguage}[2]{{%
\expandafter\ifx\csname l@#1\endcsname\relax
\typeout{** WARNING: IEEEtran.bst: No hyphenation pattern has been}%
\typeout{** loaded for the language `#1'. Using the pattern for}%
\typeout{** the default language instead.}%
\else
\language=\csname l@#1\endcsname
\fi
#2}}
\providecommand{\BIBdecl}{\relax}
\BIBdecl

\bibitem{goodfellow2014explaining}
I.~J. Goodfellow, J.~Shlens, and C.~Szegedy, ``Explaining and harnessing adversarial examples,'' \emph{arXiv preprint arXiv:1412.6572}, 2014.

\bibitem{madry2017towards}
A.~Madry, A.~Makelov, L.~Schmidt, D.~Tsipras, and A.~Vladu, ``Towards deep learning models resistant to adversarial attacks,'' \emph{arXiv preprint arXiv:1706.06083}, 2017.

\bibitem{shen2021gradient}
Y.~Shen, L.~Zheng, M.~Shu, W.~Li, T.~Goldstein, and M.~Lin, ``Gradient-free adversarial training against image corruption for learning-based steering,'' \emph{Advances in Neural Information Processing Systems}, vol.~34, 2021.

\bibitem{bhagoji2018enhancing}
A.~N. Bhagoji, D.~Cullina, C.~Sitawarin, and P.~Mittal, ``Enhancing robustness of machine learning systems via data transformations,'' in \emph{2018 52nd Annual Conference on Information Sciences and Systems (CISS)}.\hskip 1em plus 0.5em minus 0.4em\relax IEEE, 2018, pp. 1--5.

\bibitem{hendrycks2019augmix}
D.~Hendrycks, N.~Mu, E.~D. Cubuk, B.~Zoph, J.~Gilmer, and B.~Lakshminarayanan, ``Augmix: A simple data processing method to improve robustness and uncertainty,'' \emph{arXiv preprint arXiv:1912.02781}, 2019.

\bibitem{wang2021augmax}
H.~Wang, C.~Xiao, J.~Kossaifi, Z.~Yu, A.~Anandkumar, and Z.~Wang, ``Augmax: Adversarial composition of random augmentations for robust training,'' \emph{Advances in Neural Information Processing Systems}, vol.~34, 2021.

\bibitem{gong2021maxup}
C.~Gong, T.~Ren, M.~Ye, and Q.~Liu, ``Maxup: Lightweight adversarial training with data augmentation improves neural network training,'' in \emph{Proceedings of the IEEE/CVF Conference on Computer Vision and Pattern Recognition}, 2021, pp. 2474--2483.

\bibitem{shu2020prepare}
M.~Shu, Z.~Wu, M.~Goldblum, and T.~Goldstein, ``Prepare for the worst: Generalizing across domain shifts with adversarial batch normalization,'' \emph{arXiv preprint arXiv:2009.08965}, 2020.

\bibitem{heusel2017gans}
M.~Heusel, H.~Ramsauer, T.~Unterthiner, B.~Nessler, and S.~Hochreiter, ``Gans trained by a two time-scale update rule converge to a local nash equilibrium,'' \emph{Advances in neural information processing systems}, vol.~30, 2017.

\bibitem{bengio2009curriculum}
Y.~Bengio, J.~Louradour, R.~Collobert, and J.~Weston, ``Curriculum learning,'' in \emph{Proceedings of the 26th annual international conference on machine learning}, 2009, pp. 41--48.

\bibitem{ramanishka2018toward}
V.~Ramanishka, Y.-T. Chen, T.~Misu, and K.~Saenko, ``Toward driving scene understanding: A dataset for learning driver behavior and causal reasoning,'' in \emph{Proceedings of the IEEE Conference on Computer Vision and Pattern Recognition}, 2018, pp. 7699--7707.

\bibitem{sun2020scalability}
P.~Sun, H.~Kretzschmar, X.~Dotiwalla, A.~Chouard, V.~Patnaik, P.~Tsui, J.~Guo, Y.~Zhou, Y.~Chai, B.~Caine \emph{et~al.}, ``Scalability in perception for autonomous driving: Waymo open dataset,'' in \emph{Proceedings of the IEEE/CVF Conference on Computer Vision and Pattern Recognition}, 2020, pp. 2446--2454.

\bibitem{geyer2020a2d2}
J.~Geyer, Y.~Kassahun, M.~Mahmudi, X.~Ricou, R.~Durgesh, A.~S. Chung, L.~Hauswald, V.~H. Pham, M.~M{\"u}hlegg, S.~Dorn \emph{et~al.}, ``A2d2: Audi autonomous driving dataset,'' \emph{arXiv preprint arXiv:2004.06320}, 2020.

\bibitem{chen2017sully}
S.~Chen, ``Sully chen driving dataset,'' 2017.

\bibitem{bojarski2016end}
M.~Bojarski, D.~Del~Testa, D.~Dworakowski, B.~Firner, B.~Flepp, P.~Goyal, L.~D. Jackel, M.~Monfort, U.~Muller, J.~Zhang \emph{et~al.}, ``End to end learning for self-driving cars,'' \emph{arXiv preprint arXiv:1604.07316}, 2016.

\bibitem{he2016deep}
K.~He, X.~Zhang, S.~Ren, and J.~Sun, ``Deep residual learning for image recognition,'' in \emph{Proceedings of the IEEE conference on computer vision and pattern recognition}, 2016, pp. 770--778.

\bibitem{hendrycks2019benchmarking}
D.~Hendrycks and T.~Dietterich, ``Benchmarking neural network robustness to common corruptions and perturbations,'' \emph{arXiv preprint arXiv:1903.12261}, 2019.

\bibitem{roy2018robust}
S.~S. Roy, S.~I. Hossain, M.~Akhand, and K.~Murase, ``A robust system for noisy image classification combining denoising autoencoder and convolutional neural network,'' \emph{International Journal of Advanced Computer Science and Applications}, vol.~9, no.~1, pp. 224--235, 2018.

\bibitem{xiong2022robust}
Y.~Xiong and R.~Zuo, ``Robust feature extraction for geochemical anomaly recognition using a stacked convolutional denoising autoencoder,'' \emph{Mathematical Geosciences}, vol.~54, no.~3, pp. 623--644, 2022.

\bibitem{aspandi2019robust}
D.~Aspandi, O.~Martinez, F.~Sukno, and X.~Binefa, ``Robust facial alignment with internal denoising auto-encoder,'' in \emph{2019 16th Conference on Computer and Robot Vision (CRV)}.\hskip 1em plus 0.5em minus 0.4em\relax IEEE, 2019, pp. 143--150.

\bibitem{wang2020end}
T.~Wang, Y.~Luo, J.~Liu, R.~Chen, and K.~Li, ``End-to-end self-driving approach independent of irrelevant roadside objects with auto-encoder,'' \emph{IEEE Transactions on Intelligent Transportation Systems}, 2020.

\bibitem{papachristodoulou2021driveguard}
A.~Papachristodoulou, C.~Kyrkou, and T.~Theocharides, ``Driveguard: Robustification of automated driving systems with deep spatio-temporal convolutional autoencoder,'' in \emph{Proceedings of the IEEE/CVF Winter Conference on Applications of Computer Vision}, 2021, pp. 107--116.

\bibitem{xie2019feature}
C.~Xie, Y.~Wu, L.~v.~d. Maaten, A.~L. Yuille, and K.~He, ``Feature denoising for improving adversarial robustness,'' in \emph{Proceedings of the IEEE/CVF conference on computer vision and pattern recognition}, 2019, pp. 501--509.

\bibitem{liao2018defense}
F.~Liao, M.~Liang, Y.~Dong, T.~Pang, X.~Hu, and J.~Zhu, ``Defense against adversarial attacks using high-level representation guided denoiser,'' in \emph{Proceedings of the IEEE conference on computer vision and pattern recognition}, 2018, pp. 1778--1787.

\bibitem{chen2020adversarial}
T.~Chen, S.~Liu, S.~Chang, Y.~Cheng, L.~Amini, and Z.~Wang, ``Adversarial robustness: From self-supervised pre-training to fine-tuning,'' in \emph{Proceedings of the IEEE/CVF Conference on Computer Vision and Pattern Recognition}, 2020, pp. 699--708.

\bibitem{hendrycks2019using}
D.~Hendrycks, M.~Mazeika, S.~Kadavath, and D.~Song, ``Using self-supervised learning can improve model robustness and uncertainty,'' \emph{Advances in neural information processing systems}, vol.~32, 2019.

\bibitem{xu2019temporal}
M.~Xu, M.~Gao, Y.-T. Chen, L.~S. Davis, and D.~J. Crandall, ``Temporal recurrent networks for online action detection,'' in \emph{Proceedings of the IEEE/CVF International Conference on Computer Vision}, 2019, pp. 5532--5541.

\bibitem{shi2020pv}
S.~Shi, C.~Guo, L.~Jiang, Z.~Wang, J.~Shi, X.~Wang, and H.~Li, ``Pv-rcnn: Point-voxel feature set abstraction for 3d object detection,'' in \emph{Proceedings of the IEEE/CVF Conference on Computer Vision and Pattern Recognition}, 2020, pp. 10\,529--10\,538.

\bibitem{yi2021complete}
L.~Yi, B.~Gong, and T.~Funkhouser, ``Complete \& label: A domain adaptation approach to semantic segmentation of lidar point clouds,'' in \emph{Proceedings of the IEEE/CVF conference on computer vision and pattern recognition}, 2021, pp. 15\,363--15\,373.

\bibitem{kingma2014adam}
D.~P. Kingma and J.~Ba, ``Adam: A method for stochastic optimization,'' \emph{arXiv preprint arXiv:1412.6980}, 2014.

\end{thebibliography}
